\documentclass[lettersize,journal]{IEEEtran}
\usepackage{amsmath,amsfonts}
\usepackage{algorithmic}
\usepackage{algorithm}
\usepackage{array}
\usepackage[caption=false,font=footnotesize]{subfig}
\usepackage{textcomp}
\usepackage{stfloats}
\usepackage{url}
\usepackage{verbatim}
\usepackage{graphicx}
\usepackage{cite}
\usepackage[utf8]{inputenc} % allow utf-8 input
\usepackage[T1]{fontenc}    % use 8-bit T1 fonts
\usepackage{hyperref}       % hyperlinks
\usepackage{url}            % simple URL typesetting
\usepackage{booktabs}       % professional-quality tables
\usepackage{amsfonts}       % blackboard math symbols
\usepackage{nicefrac}       % compact symbols for 1/2, etc.
\usepackage{microtype}      % microtypography
\usepackage{lipsum}
\usepackage{fancyhdr}       % header
\usepackage{graphicx}
\graphicspath{{media/}} 
\usepackage{multirow}
\usepackage[table,xcdraw]{xcolor}
\usepackage{amsmath}
\usepackage{amssymb}
\begin{document}

\title{LoRA-Drop: Temporal LoRA Decoding for Efficient LLM Inference}

\author{{
  Hossein Rajabzadeh$^{1}$, Maryam Dialameh$^{1}$, Chul B. Park$^{2}$, Il-Min Kim$^{3}$, Hyock Ju Kwon$^{1}$ \\
  $^{1}$Department of Mechanical and Mechatronics Engineering, University of Waterloo\\
  $^{2}$Department of Mechanical and Industrial Engineering, University of Toronto\\ 
  $^{3}$Department of Electrical and Computer Engineering, Queen’s University \\
  \texttt{\{hossein.rajabzadeh, maryam.dialameh,hjkwon\}@uwaterloo.ca}\\ \texttt{Park@mie.utoronto.ca,Ilmin.kim@queensu.ca} 
}
        % <-this % stops a space
\thanks{}% <-this % stops a space
\thanks{}}

\markboth{}%
{Shell \MakeLowercase{\textit{et al.}}: A Sample Article Using IEEEtran.cls for IEEE Journals}

\IEEEpubid{}

\maketitle

\begin{abstract}
Autoregressive large language models (LLMs) are bottlenecked by sequential decoding, where each new token typically requires executing all transformer layers. Existing dynamic-depth and layer-skipping methods reduce this cost, but often rely on auxiliary routing mechanisms or incur accuracy degradation when bypassed layers are left uncompensated.
We present \textbf{LoRA-Drop}, a plug-and-play inference framework that accelerates decoding by applying a \emph{temporal compute schedule} to a fixed subset of intermediate layers: on most decoding steps, selected layers reuse the previous-token hidden state and apply a low-rank LoRA correction, while periodic \emph{refresh} steps execute the full model to prevent drift.
LoRA-Drop requires no routing network, is compatible with standard KV caching, and can reduce KV-cache footprint by skipping KV updates in droppable layers during LoRA steps and refreshing periodically.
Across \textbf{LLaMA2-7B}, \textbf{LLaMA3-8B}, \textbf{Qwen2.5-7B}, and \textbf{Qwen2.5-14B}, LoRA-Drop achieves up to \textbf{2.6$\times$ faster decoding} and \textbf{45--55\% KV-cache reduction} while staying within \textbf{0.5 percentage points (pp)} of baseline accuracy.
Evaluations on reasoning (GSM8K, MATH, BBH), code generation (HumanEval, MBPP), and long-context/multilingual benchmarks (LongBench, XNLI, XCOPA) identify a consistent \emph{safe zone} of scheduling configurations that preserves quality while delivering substantial efficiency gains, providing a simple path toward adaptive-capacity inference in LLMs. Codes are available at https://github.com/hosseinbv/LoRA-Drop.git. 
\end{abstract}

% \begin{IEEEkeywords}
% Article submission, IEEE, IEEEtran, journal, \LaTeX, paper, template, typesetting.
% \end{IEEEkeywords}

\section{Introduction}
\IEEEPARstart{L}{arge} Language Models (LLMs) have emerged as a cornerstone of modern artificial intelligence, demonstrating remarkable performance across a wide range of natural language processing tasks such as reasoning, code generation, and dialogue systems \cite{brown2020language,chowdhery2023palm,openai2023gpt}. Their ability to generalize across domains with minimal task-specific supervision has driven widespread adoption in both academia and industry, powering applications in search engines, conversational agents, recommendation systems, and enterprise productivity tools \cite{thoppilan2022lamda,bubeck2023sparks,team2023gemini}.

Despite their success, the practical deployment of LLMs is often hindered by the high computational cost of autoregressive inference, where each token must be generated sequentially through the entire stack of transformer layers. This process leads to substantial latency and energy consumption, especially for long sequences and interactive applications. To address this challenge, several lines of research have emerged:

\textbf{Model Compression and Quantization}-
Methods such as weight pruning \cite{frantar2023sparsegpt,ma2023llm} and low-bit quantization \cite{dettmers2023qlora,xiao2023smoothquant, rajabzadeh2024qdylora} reduce the parameter footprint and arithmetic cost of LLMs, enabling more efficient deployment on constrained hardware. While effective for reducing memory and throughput requirements, these approaches often require delicate tuning to balance efficiency with model accuracy.

\begin{figure}[t]
    \centering
    \includegraphics[width=\linewidth]{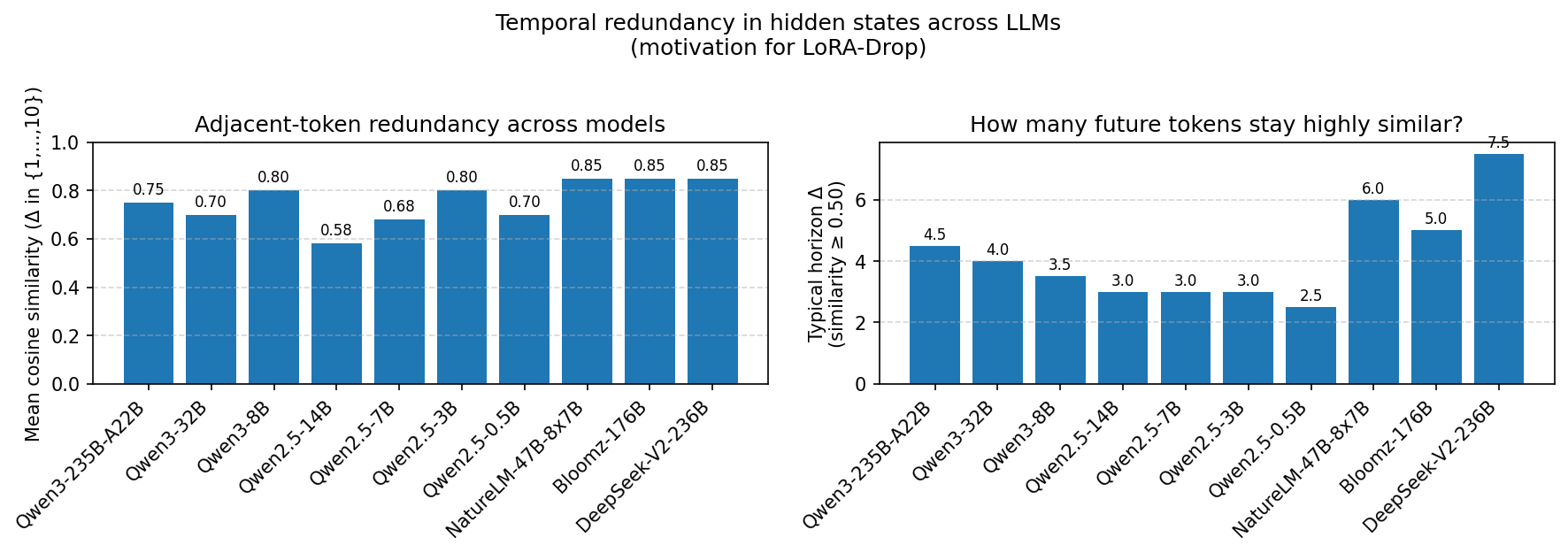}
    \caption{
\textbf{(Left)} Adjacent-token redundancy measured as the average cosine similarity between the hidden state of token $t$ and those of the next tokens $t+\Delta$ (with $\Delta \in \{1,...,10\}$), averaged over all positions $t$ across 1024 batches of diverse data. All evaluated models show high similarity (approximately $0.6$--$0.85$), indicating that hidden states change little between consecutive tokens. \textbf{(Right)} The \emph{similarity horizon}, defined as the largest
    token distance $\Delta$ for which the cosine similarity between token $t$
    and $t+\Delta$ remains at least $0.50$. Horizons between $3$ and $6$
    tokens demonstrate that several future tokens share highly similar hidden
    states.
    }
    \label{fig:temporal_redundancy}
    \vspace{-16pt} % <-- removes space AFTER caption
\end{figure}

\textbf{Efficient Attention and KV-Caching-} Given that attention layers dominate LLM inference cost, many works focus on optimizing sequence processing. KV-caching \cite{korthikanti2023reducing,dialameh2025echo,rajabzadeh2024echoattattendcopyadjust} reduces redundant computation across tokens, while efficient attention variants such as FlashAttention \cite{dao2022flashattention} and streaming/context-parallel attention \cite{dao2023flashattention2,zheng2024ringattention} accelerate long-context inference. These methods improve throughput but do not directly address the redundancy across layers during generation.

\textbf{Dynamic Computation and Layer Skipping-} More recently, researchers have proposed adaptive computation strategies that reduce the number of active layers or sublayers per token. Examples include Unified Layer Skipping \cite{liu2024unified}, FlexiDepth \cite{luo2025flexidepth}, AdaSkip \cite{he2025adaskip}, ColT5 \cite{ainslie2023colt5}, Balcony \cite{jamialahmadi2025balcony}, and FiRST \cite{jain2024first}, which either skip intermediate layers or use lightweight routers and adapters to allocate variable computational budgets across tokens. These methods highlight that not all tokens require the full depth of the model, though many approaches either incur additional routing overhead or bypass layers entirely without compensatory transformations, leading to potential degradation.

Recent interpretability studies have shown that the intermediate layer representations of large language models (LLMs) are remarkably expressive, often containing sufficient information to anticipate not only the next token but even subsequent ones. Logit Lens \cite{nostalgebraist2020logitlens,wang2025logitlens4llms} first, and recently demonstrated that hidden states from early and middle layers already encode meaningful token distributions when projected through the model’s output embedding, suggesting that prediction refinement, rather than new information synthesis, occurs as tokens propagate through deeper layers. Building on this, Tuned Lens \cite{belrose2023eliciting} aligned each layer’s representation to the output space via lightweight linear transformations, revealing that layer outputs form a coherent progression of increasingly confident latent predictions. Extending this temporal perspective, Future Lens \cite{pal2023future} showed that a single hidden state can often anticipate multiple future tokens, indicating significant temporal redundancy across decoding steps. Together, these works suggest that the intermediate activations of autoregressive transformers already encapsulate much of the predictive signal used in later computations, implying that full-depth inference at every time step is not always necessary.

Following the reported representation similarities across layers in LLMs, our analysis reveals that large language models exhibit a striking degree of temporal redundancy in their hidden states. As shown in Fig.~\ref{fig:temporal_redundancy}, adjacent tokens in models such as Qwen, DeepSeek, and NatureLM maintain very high cosine similarity (0.6–0.85 on average), and this similarity persists across several future positions—typically up to 3–6 tokens. This indicates that the model's internal representation at time step $t$ already contains much of the information needed for the representations at steps $t+1,\dots,t+\Delta$ \footnote{$\Delta$ denotes the temporal offset between two token positions,
i.e., the number of future nearby tokens over which hidden-state similarity
is evaluated in the LoRA-Drop analysis.}, even before these tokens are generated. Building on this observation, we propose LoRA-Drop, a method that directly exploits the inherent predictability of hidden states to reduce unnecessary computation during autoregressive decoding. 

Whereas existing dynamic inference methods rely on complex routing or skip layers without compensatory updates—often degrading generation quality—we paper propose LoRA-Drop which reuses the rich intermediate representation from step $t$ and applies a lightweight LoRA transformation at step $t+\Delta$ to adjust it for the next tokens. This design preserves contextual continuity while avoiding full-layer computation at every step. LoRA-Drop offers two key benefits. First, it enables fine-grained control over the drop ratio, allowing practitioners to decide how frequently the full model should be invoked versus the fast LoRA-only pathway, thereby flexibly trading off latency and accuracy. Second, it is fully plug-and-play: it introduces no architectural changes, integrates seamlessly with pretrained LLMs, and requires only a small amount of continual fine-tuning.

% ============================================================
\section{Proposed Method: LoRA-Drop}
\label{sec:lora-drop}
% ============================================================

\begin{figure*}[t]
    \centering
    \includegraphics[width=0.7\textwidth]{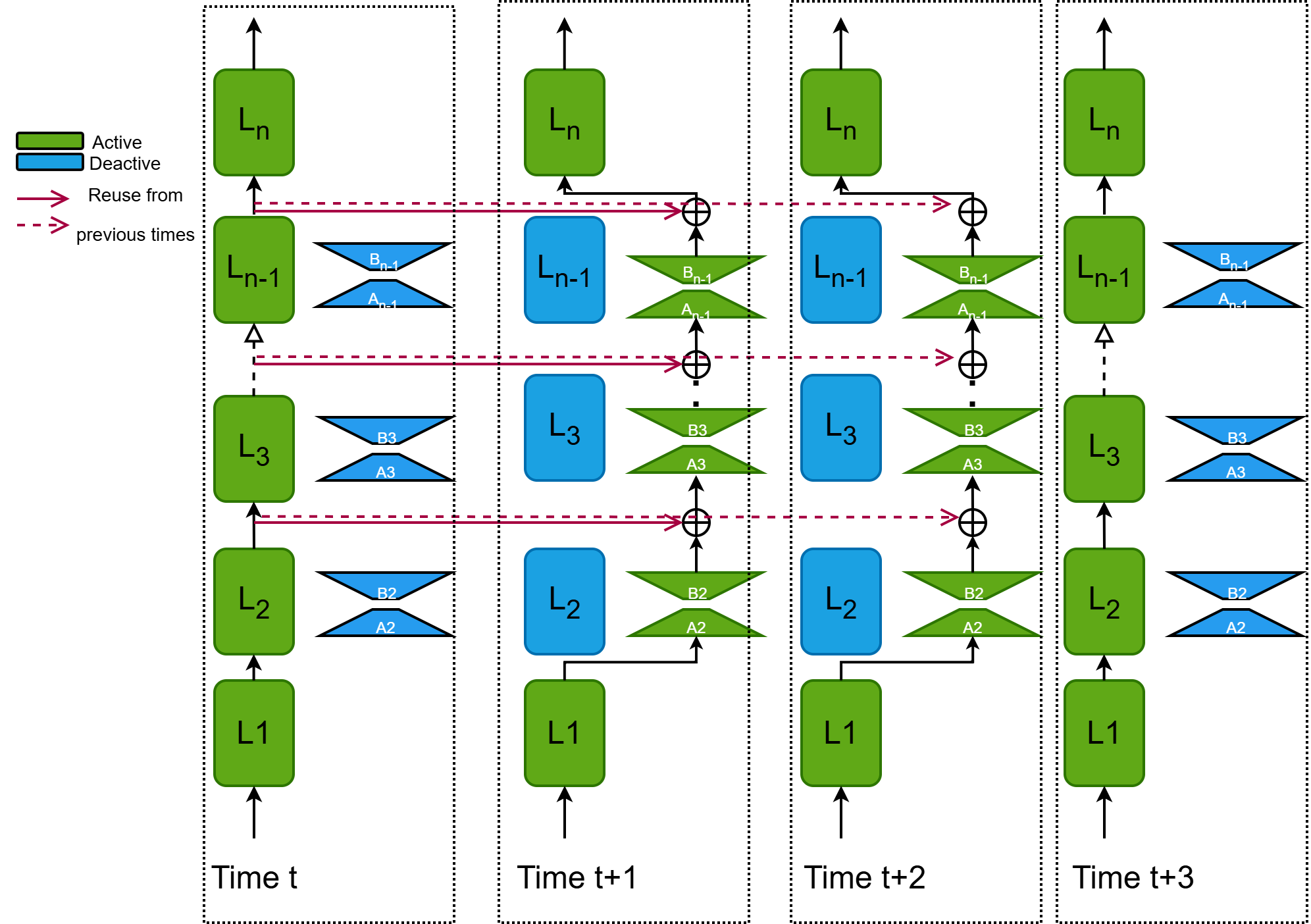}
    \caption{The workflow of \textbf{LoRA-Drop}. Each transformer layer is augmented with a low-rank adaptation module (matrices $A_i$ and $B_i$). During low-complexity steps (eg. $t+1$ and $t+2$), only the lightweight LoRA modules are activated, bypassing the pre-specified layers to reduce inference computation. At periodic or complexity-triggered steps, the full layers are reactivated to refine representations.}
    \label{fig:lora-drop}
\end{figure*}

The goal of \textbf{LoRA-Drop} is to accelerate the inference process of autoregressive in large language models by dynamically modulating the model’s computational capacity according to the initial layers' similarities. Instead of computing all transformer layers at every time step, LoRA-Drop allows certain layers to be \emph{skipped} while preserving representational continuity through lightweight low-rank adaptation (LoRA) modules. Figure \ref{fig:lora-drop} illustrates the main workflow of LoRA-Drop using an inference example with a sequence length of four. As shown in the figure, at time step $t$, the model performs a standard forward pass through all layers, while the LoRA modules are disabled.
At the subsequent time step $t+1$, a subset of intermediate layers is skipped: instead of executing their full computations, the corresponding LoRA modules are activated, and their outputs are added to the cached outputs of the same layers from the previous time step. The same computational workflow is repeated at time step $t+2$, and a full-layer computation is performed again at time step $t+3$, allowing the model to periodically refine its representations across all layers. This produces an approximate update of the hidden states while avoiding the full layer computation.
This approach leverages the empirical observation that intermediate layer representations of LLMs already encode rich predictive information about upcoming tokens \cite{nostalgebraist2020logitlens}. Thus, the full model call is not always required for predicting nearby tokens.

% -------------------------------
% -------------------------------

\subsection{Model Formulation}

Consider an $n$-layer autoregressive transformer receiving an input sequence of tokens $S=\{t_1, t_2, \dots, t_T\}$ and generating a sequence of $m$ tokens $S^{\prime}=\{\hat{t}_{T+1}, \hat{t}_{T+2}, \dots, \hat{t}_{T+m}\}$. Let $x_t^i \in \mathbb{R}^{d}$ denote the output hidden state of layer $i$ and time step $t$, where $d$ is the model's dimension. The standard forward propagation for layer $i$ is given by
\begin{equation}
x_{t}^i = f^i(x_t^{i-1}),
\label{eq:standard-layer}
\end{equation}
where $f^i(\cdot)$ represents the standard transformation performed by the attention, normalization, and MLP modules of layer $i$ and $x_t^{i-1}$ is the output of layer $i-1$ at time step $t$. 

In LoRA-Drop, we first need to define a list of drop-layers, i.e. $\mathbb{L}$, \footnote{Please check Subsection \ref{drop-layer}, to see how the list of drop-layers is created.}, in which each layer $i \in \mathbb{L}$ in that list are equipped with a \textit{LoRA module} consisting of two learnable low-rank matrices $A_i \in \mathbb{R}^{r \times d}$ and $B_i \in \mathbb{R}^{d \times r}$, with rank $r \ll d$. These matrices approximate a residual transformation of the form
\begin{equation}
W_L^i = B_i A_i, \qquad \text{rank}(W_L^i) = r.
\end{equation}
During selected time steps, the main layer computation $f^i(\cdot)$ is bypassed and replaced by this lightweight linear mapping:
\begin{equation}
\hat{x}_{t}^i = x_{t-1}^i + \alpha \, W_L^i x_t^{i-1},
\label{eq:lora-drop}
\end{equation}
where $\alpha$ is a scaling coefficient controlling the contribution of the LoRA update. Equation~\eqref{eq:lora-drop} serves as a compressed surrogate for the full layer output in~\eqref{eq:standard-layer}.
\subsection{Temporal Scheduling of Layer Activation}
\label{sec:temporal_scheduling}

LoRA-Drop accelerates autoregressive decoding by temporally scheduling a subset of
transformer layers to alternate between (i) a full evaluation and (ii) a lightweight
LoRA-only surrogate update. Let the model have $n$ layers and let
$\mathbb{L}\subseteq\{1,\dots,n\}$ denote the fixed \emph{drop-layer list}.
We define $\rho \triangleq |\mathbb{L}|/n$ as the fraction of layers that are
\emph{droppable} (i.e., eligible for LoRA-mode). Layers outside $\mathbb{L}$ are
always executed in full.

\paragraph{Two decoding modes and refresh period.}
Decoding proceeds in cycles of length $k{+}1$:
\begin{itemize}
    \item \textbf{Full mode (refresh step).} Every $(k{+}1)$-th decoding step,
    all layers are executed in full. This \emph{refresh} recomputes exact hidden
    states and updates internal states such as KV-cache, preventing drift from
    accumulating over long generations.

    \item \textbf{LoRA mode (lightweight steps).} For the following $k$ decoding
    steps, layers in $\mathbb{L}$ bypass the full computation and instead apply
    a low-rank LoRA update with cost $\mathcal{O}(rd)$, while layers not in
    $\mathbb{L}$ remain fully active. In this phase, LoRA-Drop exploits temporal
    redundancy in hidden states across nearby tokens to approximate the effect of
    the skipped computation.
\end{itemize}
In this work, we adopt the above \emph{fixed periodic schedule}; we leave adaptive
confidence-based scheduling (e.g., via token entropy or logit margin) as future work.

\paragraph{Activation indicator with $(\rho,k)$ control.}
Let $\delta_t^i\in\{0,1\}$ indicate whether layer $i$ is executed in full at decoding
step $t$:
\[
\delta_t^i =
\begin{cases}
1, & \text{if } (t \bmod (k{+}1)=0)\ \ \text{(refresh step)},\\
1, & \text{if } i \notin \mathbb{L}\ \ \text{(non-droppable layer)},\\
0, & \text{otherwise (LoRA-mode on droppable layers).}
\end{cases}
\]
By construction, exactly a $\rho$ fraction of layers can take $\delta_t^i=0$ on
non-refresh steps, and $k$ controls how many consecutive tokens are generated before
the next full refresh.

\paragraph{Unified layer update.}
Let $x_t^{i-1}$ be the input activation to layer $i$ at token step $t$, and let
$f^i(\cdot)$ denote the original full layer transformation. The LoRA-Drop update is
\begin{equation}
x_t^i
=
\delta_t^i\, f^i(x_t^{i-1})
+
(1-\delta_t^i)\left(
x_{t-1}^{i} + \alpha\, W_L^i\, x_t^{i-1}
\right),
\label{eq:unified-update}
\end{equation}
where $x_t^i$ is the layer output at time $t$. In LoRA mode ($\delta_t^i=0$),
the layer output is approximated by reusing the cached hidden state from the previous
token at the same layer, $x_{t-1}^i$, plus a low-rank LoRA correction applied to the
current layer input $x_t^{i-1}$. On refresh steps and for non-droppable layers
($\delta_t^i=1$), LoRA-Drop reduces to the standard full computation.

% \subsection{Temporal Scheduling of Layer Activation}

% The inference process alternates between two computational modes:

% \begin{itemize}
%     \item \textbf{LoRA mode (lightweight phase):} for low-complexity tokens or pre-specified steps, only LoRA modules are activated; the main layers are bypassed. This yields a fast linear transformation with cost $\mathcal{O}(r d)$.
%     \item \textbf{Full mode (refinement phase):} every $k$ decoding steps (or based on a learned gating signal), all layers $f^i(\cdot)$ are reactivated to restore full model capacity and update contextual representations.
% \end{itemize}

% The switching policy between these modes can follow a fixed schedule or an adaptive function based on token confidence metrics \footnote{In this work, we follow the the fixed schedule strategy and leave the confidence based strategy as future work.} (e.g., the probability of previously generated token). Denoting the activation indicator of layer $i$ at time $t$ as
% \[
% \delta_t^i =
% \begin{cases}
% 1, & \text{if full layer } L_i \text{ is active},\\
% 0, & \text{if only LoRA module is active},
% \end{cases}
% \]
% the unified update rule becomes
% \begin{equation}
% x_{t}^i = \delta_t^i \, f^i(x_t^{i-1}) + (1 - \delta_t^i)\left(x_t^{i-1} + \alpha W_L^i x_t^{i-1}\right).
% \label{eq:unified-update}
% \end{equation}

% where $x^i_t$ is the final output of layer $i$ for output token at time $t$, and $f^i(x_t^{i-1})$ is the immediate output of layer $i$ for the input of $x_t^{i-1}$.
\subsection{LoRA-Drop Inference Algorithm}

Algorithm~\ref{alg:lora-drop} formalizes the decoding process under LoRA-Drop inference strategy. Each new token generation step decides whether to invoke the full layer computation or its LoRA surrogate according to the drop-layer list $\mathbb{L}$ and the given activation period $k\in{1,...,\Delta}$.
As explained in the algorithm, LoRA-Drop alternates between periodic full-model refresh steps and lightweight LoRA-only updates: at every decoding step $j$, if $j \bmod k=0$, all layers execute a standard forward pass to refresh hidden states and KV-caches, whereas for intermediate steps, layers whose indices belong to the drop-layer list
$\mathbb{L}$ apply a low-rank LoRA update based on the previous hidden state, while the remaining layers continue to perform full forward computation. Specifically, decoding proceeds token by token. For each newly generated token position $j$, the model iterates through all transformer layers in order. If the current step corresponds to a refresh point ($j \bmod k =0$), the full network is evaluated, ensuring that all hidden representations and internal states are recomputed exactly. Otherwise, for layers included in $\mathbb{L}$, the computation is approximated using a LoRA-only residual update added to the cached hidden state from the previous token, while layers outside $\mathbb{L}$ continue to execute their standard forward functions. The resulting top-layer representation is then used to predict the next token, and the process repeats until the desired output length is reached. This approach preserves compatibility with KV caching since both update paths maintain consistent hidden-state dimensionality.

% -------------------------------
\begin{algorithm}[h]
\caption{LoRA-Drop Inference}
\label{alg:lora-drop}
\begin{algorithmic}[1]
\STATE \textbf{Input:} Token sequence $S=\{t_1, \dots, t_T\}$, model layers $\{L_i\}_{i=1}^n$, LoRA modules $\{W_L^i\}_{i=1}^n$, activation period $k$, Drop-layer list $\mathbb{L}$
\STATE \textbf{Output:} Generated m-token sequence $\{\hat{t}_{T+1},\dots\,\hat{t}_{T+m}\}$ 
\FOR{$j \in \{{T+1},..., T+m$\}}
    \FOR{each layer $L_i : i = 1, 2, \dots, n$}
    \IF{$j \bmod k = 0$}
        \STATE Full forward pass with all layers are active: $x_{j}^i = f^i(x_{j}^{i-1})$
        \ELSIF{$i \in \mathbb{L}$}
            \STATE LoRA mode: $x_{j}^i = x_{j-1}^{i} + \alpha W_L^i x_j^{i-1}$
        \ELSE
            \STATE 
            Full forward pass with all layers are active: $x_{j}^i = f^i(x_j^{i-1})$
        \ENDIF
    \ENDFOR
    \STATE Predict next the token $\hat{t}_{j}$
\ENDFOR
\end{algorithmic}
\end{algorithm}
% -------------------------------

\subsection{Computational Analysis}
\label{sec:comp_analysis}

We analyze LoRA-Drop during \emph{autoregressive decoding}. Let the model have $n$
transformer layers and let $\mathbb{L}$ be the set of droppable layers with
$|\mathbb{L}|=\rho n$, where $\rho\in[0,1]$ is the fraction of droppable layers.
LoRA-Drop performs an exact \emph{refresh} step once every $k{+}1$ tokens; the
remaining $k$ steps use LoRA-mode in layers $\mathbb{L}$ while always-on layers
remain fully active.

\paragraph{Decode-time layer cost decomposes into projection + attention-to-cache.}
Let $L$ denote the KV-cache length at a given decode step (prompt length plus
generated tokens so far). For a standard transformer layer at decode time, the
dominant work consists of: (i) dense projections and MLP blocks (typically
$\Theta(d^2)$), and (ii) attending the current query to the cached keys/values
(typically $\Theta(dL)$). We therefore write the \emph{per-layer} full cost as
\begin{equation}
C_{\text{full}}(L) \;=\; A d^2 \;+\; B d L,
\label{eq:cfull}
\end{equation}
where constants $A,B>0$ absorb architectural factors (heads, MLP expansion,
implementation details, etc.). Crucially, $C_{\text{full}}(L)$ grows linearly in
cache length $L$ via the attention-to-cache term.

In LoRA-mode, a droppable layer replaces the full computation with a low-rank
surrogate update, whose dominant cost is the LoRA matvec/matmul:
\begin{equation}
C_{\text{LoRA}} \;=\; \Theta(rd), \qquad r\ll d,
\label{eq:clora}
\end{equation}
which is \emph{independent} of $L$ (it does not attend to the cache and does not
compute full projections/MLP for that layer).

\paragraph{Cycle-average per-token compute.}
Let $C_{\text{base}}(L)=n\,C_{\text{full}}(L)$ be the baseline per-token decode cost
(all layers full). Under LoRA-Drop, each period of length $k{+}1$ contains one
refresh step (all layers full) and $k$ LoRA-steps (droppable layers use LoRA-mode).
Thus the \emph{cycle-average} per-token cost is
{\small
\begin{align}
C_{\text{avg}}(L)
&=\underbrace{(1-\rho)n\,C_{\text{full}}(L)}_{\text{always-on layers}} \;+\;
\underbrace{\rho n\left(\frac{1}{k+1}C_{\text{full}}(L) + \frac{k}{k+1}C_{\text{LoRA}}\right)}_{\text{droppable layers}} \nonumber\\
&= n\,C_{\text{full}}(L)\left[(1-\rho)+\frac{\rho}{k+1}+\frac{\rho k}{k+1}\,\gamma(L)\right],
\label{eq:cavg}
\end{align}
where we define the \emph{relative surrogate ratio}
\begin{equation}
\gamma(L)\;\triangleq\;\frac{C_{\text{LoRA}}}{C_{\text{full}}(L)}
\;=\;\Theta\!\left(\frac{r}{A d + B L}\right).
\label{eq:gammaL}
\end{equation}
}
Equation~\eqref{eq:gammaL} shows that $\gamma(L)$ \emph{decreases} as $L$ grows:
LoRA-mode becomes relatively cheaper for longer contexts because it eliminates
the $\Theta(dL)$ attention-to-cache work in the dropped layers.

\paragraph{Closed-form speedup with long-context scaling.}
Define the idealized compute speedup as
$S(L)\triangleq C_{\text{base}}(L)/C_{\text{avg}}(L)$. Using~\eqref{eq:cavg} we obtain
\begin{equation}
S(L)
=
\frac{1}{
(1-\rho)+\frac{\rho}{k+1}+\frac{\rho k}{k+1}\,\gamma(L)
}.
\label{eq:speedupL}
\end{equation}

\textbf{Meaningful simplified cases.}
\begin{itemize}
\item \textbf{Long-context limit ($L\to\infty$).}
Since $\gamma(L)\to 0$ as $L$ increases, the speedup approaches
\begin{equation}
S_{\infty}
\;\triangleq\;\lim_{L\to\infty} S(L)
=
\frac{1}{(1-\rho)+\frac{\rho}{k+1}}
=
\frac{k+1}{(k+1)-\rho k}.
\label{eq:speedup_long_context}
\end{equation}
This yields a tight, interpretable ceiling controlled only by $(\rho,k)$.

\item \textbf{Short-context / projection-dominated regime.}
When $L$ is small, $C_{\text{full}}(L)\approx A d^2$ and
$\gamma(L)\approx \Theta(r/d)$, recovering the familiar low-rank ratio.

\item \textbf{Diminishing returns in $k$.}
From~\eqref{eq:speedupL}, the marginal gain of increasing $k$ decays as
$\Theta(1/(k+1)^2)$ (holding $\rho$ and $\gamma(L)$ fixed), i.e., gains saturate
quickly once refresh overhead $\rho/(k+1)$ becomes small.
\end{itemize}

\paragraph{Tail-latency characterization (p95 token latency).}
LoRA-Drop induces a periodic bimodal per-token latency: refresh tokens (every
$k{+}1$ steps) are slower than LoRA-steps. Let $\tau_{\text{ref}}(L)$ and
$\tau_{\text{lora}}(L)$ denote the per-token wall-clock time of refresh and
LoRA-steps, respectively (each includes all layers for that step). Then the
fraction of slow tokens is exactly $1/(k+1)$, implying the token-latency quantiles
are determined by this frequency. In particular,
\begin{equation}
\tau_{p}(L)=
\begin{cases}
\tau_{\text{ref}}(L), & \text{if } \frac{1}{k+1} > 1-p,\\
\tau_{\text{lora}}(L), & \text{otherwise.}
\end{cases}
\label{eq:pquantile}
\end{equation}
For example, the 95th-percentile token latency satisfies
$\tau_{0.95}(L)=\tau_{\text{ref}}(L)$ whenever $k<19$, and
$\tau_{0.95}(L)=\tau_{\text{lora}}(L)$ when $k\ge 19$.
This makes explicit how $k$ controls not only average throughput but also
tail latency in serving settings.

% \subsection{Computational Analysis}

% Let $C_{\text{full}}$ denote the cost of a standard transformer layer and $C_{\text{LoRA}}$ the cost of a LoRA module. For rank $r \ll d$, we have $C_{\text{LoRA}} \approx \mathcal{O}(r d)$ and $C_{\text{full}} \approx \mathcal{O}(d^2)$. Over a decoding sequence of $T$ tokens with activation ratio $\rho$ (fraction of steps where full layers are active), the expected per-token cost becomes
% \begin{equation}
% \mathbb{E}[C_{\text{LoRA-Drop}}] = \rho \, C_{\text{full}} + (1 - \rho) \, C_{\text{LoRA}}.
% \end{equation}
% Hence, the theoretical speedup scales approximately as
% \begin{equation}
% S \approx \frac{C_{\text{full}}}{\rho \, C_{\text{full}} + (1 - \rho) \, C_{\text{LoRA}}}.
% \end{equation}
% In practice, since LoRA modules contain orders-of-magnitude fewer parameters than the original layer weights, significant reductions in inference latency can be achieved without retraining the backbone network. 

\subsection{Integration and Fine-Tuning}

LoRA-Drop can be seamlessly integrated into existing pretrained models by inserting LoRA modules in the desired subset of layers and performing a few rounds of continual fine-tuning on a small corpus. This process adapts the low-rank parameters $\{A_i, B_i\}$ while freezing the original model weights, preserving pretrained knowledge. The method thus enables \emph{post-hoc acceleration} of any transformer-based LLM without altering its architecture or requiring full retraining.

% ============================================================
\section{Experiments}
\label{sec:experiments}
% ============================================================

We evaluate the effectiveness of \textbf{LoRA-Drop} across multiple open-weight large language models (LLMs) and standard reasoning, knowledge, and commonsense benchmarks. Our experiments aim to answer three key questions:  
(1) Can LoRA-Drop accelerate inference while preserving accuracy across diverse model families and sizes?  
(2) How does the selective activation of full layers affect performance on complex versus simple reasoning tasks?  
(3) What trade-offs emerge between drop ratio, latency, and accuracy?

\subsection{Experimental Setup}

\paragraph{Models.}
We apply LoRA-Drop to four widely used autoregressive LLMs: \textbf{LLaMA 2-7B} \cite{touvron2023llama}, \textbf{LLaMA 3-8B} \cite{dubey2024llama}, \textbf{Qwen 2.5-7B}, and \textbf{Qwen 2.5-14B} \cite{yang2025qwen2}.  
Each model is equipped with LoRA modules inserted in all intermediate transformer blocks, following the formulation in Section~\ref{sec:lora-drop}.  
For stability and domain generalization, the LoRA-Drop versions of these models underwent a short stage of \emph{continual pretraining} over approximately 15 billion tokens drawn from the publicly available \textbf{RefinedWeb} corpus \cite{penedo2024the}.  
During this phase, only the LoRA parameters $\{A_i, B_i\}$ were updated, while the original model weights were frozen.

\paragraph{Datasets.}
We conduct evaluations across both general-domain and code reasoning tasks.  
For general natural language understanding, we employ the \textbf{LM-Eval Harness} suite \cite{eval-harness}, covering:
\textbf{MMLU} (multi-task knowledge), \textbf{HellaSwag} (HS: commonsense reasoning), \textbf{WinoGrande} (coreference), \textbf{ARC-c} and \textbf{ARC-e} (science QA), \textbf{OpenBookQA} (OB), \textbf{PIQA} (physical reasoning), and \textbf{RACE} (reading comprehension; denoted as RA).  
To assess generative reasoning, we additionally evaluate on the \textbf{HumanEval} dataset \cite{chen2021evaluating}, which measures code synthesis accuracy under strict functional correctness.

\paragraph{Baselines.}
Each LoRA-Drop variant is compared against its original full model (without layer skipping or LoRA modules) and against dynamic-depth baselines such as \textbf{Unified Layer Skipping} \cite{liu2024unified} and \textbf{FlexiDepth} \cite{luo2025flexidepth}.  
We report both performance metrics (accuracy, Pass@1) and efficiency metrics (tokens/sec, relative FLOPs, and memory footprint). We focus our empirical comparisons on the closest methodological baselines—i.e., depth/activation scheduling via layer skipping (e.g., Unified Layer Skipping and FlexiDepth)—because they share the same core knob as LoRA-Drop: selectively reducing per-token computation by deactivating a subset of layers during decoding. In contrast, other acceleration families such as speculative decoding, quantization, and KV-cache eviction/compression are largely orthogonal to our design and introduce additional confounding factors (e.g., draft-model selection and acceptance criteria, quantization calibration and kernel availability, or cache-management heuristics and memory allocators) that can dominate results unless each method is extensively tuned under the same serving stack and hardware-specific kernels. A fair, apples-to-apples comparison would therefore require substantial engineering and hyperparameter sweeps across multiple systems implementations, which is outside the scope of this work. Importantly, these methods are complementary to LoRA-Drop and can be combined; we leave systematic cross-family benchmarking and composition studies to future work.

\paragraph{Implementation Details.}
All experiments are performed on NVIDIA A100 and V100 GPUs with Scaled Dot-Product Attention (SDPA) and mixed-precision (BF16) inference.  
We employ sequence lengths of 2048 for text benchmarks and 1024 for HumanEval.  
Each model uses a fixed LoRA rank $r = 16$ and scaling $\alpha = 16$, unless otherwise stated.  
We vary the drop ratio $\rho \in \{0.25, 0.5, 0.75\}$ to control how frequently the full layers are activated, following the unified update rule in Equation~\eqref{eq:unified-update}.  
The LM-Eval Harness version 0.4.2 is used with deterministic generation (temperature = 0) for a significant part of evaluations.  

\begin{table*}[t]
\centering
\caption{
Evaluation of \textbf{LoRA-Drop}, \textbf{Unified Layer Skipping}~\cite{liu2024unified}, and \textbf{FlexiDepth}~\cite{luo2025flexidepth} under temporal layer skipping. 
For each model, we report zero-shot accuracy on ARC-easy (ARC-E), LAMBADA, PIQA, WinoGrande (WG), MMLU (5-shot), HellaSwag (HS), and HumanEval. 
LoRA-Drop operates by skipping a subset of intermediate layers for the next $k=3$ generated tokens: 
for a drop ratio of $\rho$=0.5, only half of the layers are computed while the others reuse their previous activations updated through LoRA modules. 
The first three and last layers remain active across all steps. 
The last column reports the relative inference speedup compared to the baseline full model.
}
\label{tab:lora-drop-eval}
\resizebox{\textwidth}{!}{
\begin{tabular}{lccccccccc}
\toprule
\textbf{Model} & \textbf{ARC-E} & \textbf{LAMBADA} & \textbf{PIQA} & \textbf{WinoG.} & \textbf{MMLU (5)} & \textbf{HS} & \textbf{HumanEval} & \textbf{Avg.} & \textbf{Speedup (×)} \\ 
\midrule

\multicolumn{10}{l}{\textbf{LLaMA2-7B}} \\ 
Full (baseline) & 75.2 & 68.2 & 78.8 & 69.2 & 45.3 & 77.6 & 38.1 & 64.6 & 1.00 \\
Unified Layer Skipping & 74.3 & 67.1 & 77.9 & 68.4 & 44.5 & 76.5 & 37.6 & 63.8 & 1.42 \\
FlexiDepth & 74.8 & 67.6 & 78.3 & 68.7 & 44.9 & 77.1 & 37.8 & 64.2 & 1.55 \\
LoRA-Drop (25\%) & 75.3 & 68.4 & 78.9 & 69.3 & 45.6 & 77.8 & 38.2 & 64.8 & 1.37 \\
LoRA-Drop (50\%) & 75.0 & 68.1 & 78.6 & 69.0 & 45.2 & 77.4 & 38.0 & 64.5 & 1.68 \\
LoRA-Drop (75\%) & 73.4 & 66.7 & 76.9 & 67.5 & 43.1 & 75.2 & 37.1 & 62.8 & 2.35 \\ 
\midrule

\multicolumn{10}{l}{\textbf{Qwen2.5-7B}} \\ 
Full (baseline) & 77.1 & 70.5 & 79.8 & 71.4 & 47.0 & 79.9 & 39.5 & 66.5 & 1.00 \\
Unified Layer Skipping & 76.3 & 69.7 & 79.0 & 70.7 & 46.2 & 79.0 & 39.0 & 65.7 & 1.38 \\
FlexiDepth & 76.7 & 70.0 & 79.3 & 70.9 & 46.5 & 79.3 & 39.2 & 66.0 & 1.52 \\
LoRA-Drop (25\%) & 77.2 & 70.6 & 79.7 & 71.6 & 47.2 & 80.0 & 39.5 & 66.5 & 1.39 \\
LoRA-Drop (50\%) & 77.0 & 70.3 & 79.5 & 71.3 & 46.9 & 79.7 & 39.3 & 66.3 & 1.73 \\
LoRA-Drop (75\%) & 74.9 & 68.2 & 77.3 & 69.1 & 44.5 & 77.0 & 37.9 & 64.1 & 2.42 \\ 
\midrule

\multicolumn{10}{l}{\textbf{LLaMA3-8B}} \\ 
Full (baseline) & 78.0 & 72.6 & 80.3 & 73.8 & 48.1 & 80.4 & 40.7 & 67.7 & 1.00 \\
Unified Layer Skipping & 77.2 & 71.7 & 79.5 & 73.0 & 47.2 & 79.5 & 40.0 & 66.9 & 1.36 \\
FlexiDepth & 77.5 & 72.1 & 79.8 & 73.3 & 47.6 & 79.8 & 40.3 & 67.2 & 1.50 \\
LoRA-Drop (25\%) & 78.1 & 72.7 & 80.4 & 73.8 & 48.3 & 80.5 & 40.8 & 67.8 & 1.34 \\
LoRA-Drop (50\%) & 77.9 & 72.4 & 80.1 & 73.6 & 48.0 & 80.2 & 40.5 & 67.5 & 1.70 \\
LoRA-Drop (75\%) & 75.8 & 70.2 & 77.8 & 71.4 & 45.5 & 77.4 & 39.1 & 65.3 & 2.38 \\ 
\midrule

\multicolumn{10}{l}{\textbf{Qwen2.5-14B}} \\ 
Full (baseline) & 80.1 & 74.8 & 81.0 & 75.5 & 50.2 & 81.2 & 42.3 & 69.3 & 1.00 \\
Unified Layer Skipping & 79.3 & 74.0 & 80.2 & 74.7 & 49.3 & 80.3 & 41.8 & 68.5 & 1.34 \\
FlexiDepth & 79.6 & 74.3 & 80.5 & 75.0 & 49.6 & 80.6 & 42.0 & 68.8 & 1.49 \\
LoRA-Drop (25\%) & 80.2 & 74.9 & 81.1 & 75.6 & 50.3 & 81.3 & 42.4 & 69.4 & 1.35 \\
LoRA-Drop (50\%) & 80.0 & 74.6 & 80.9 & 75.4 & 50.1 & 81.0 & 42.2 & 69.1 & 1.68 \\
LoRA-Drop (75\%) & 77.8 & 72.1 & 78.6 & 73.0 & 47.2 & 78.3 & 40.5 & 66.8 & 2.60 \\ 
\bottomrule
\end{tabular}
}
\end{table*}

\begin{figure*}[t]
\centering
\subfloat[LLaMA2-7B]{%
  \includegraphics[width=0.485\textwidth]{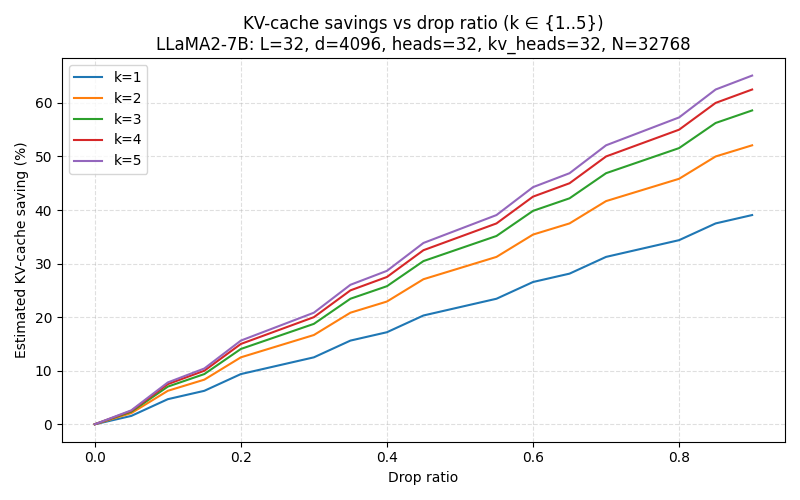}%
}\hfill%
\subfloat[Qwen2.5-7B]{%
  \includegraphics[width=0.485\textwidth]{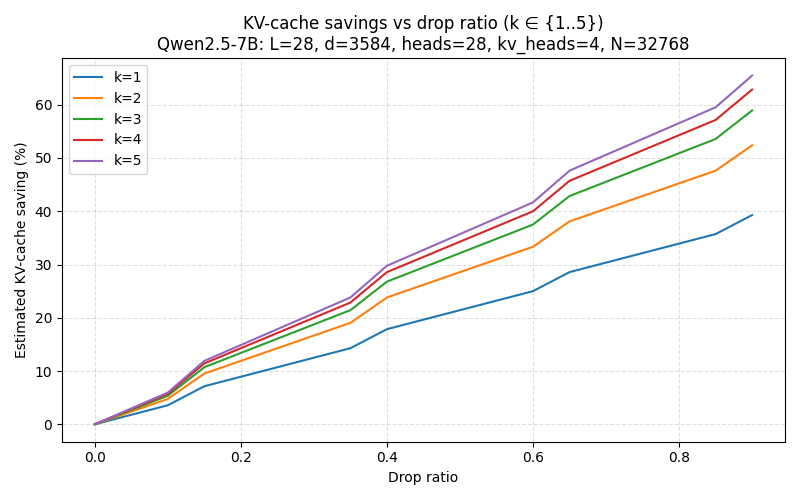}%
}\\[0.6em]
\subfloat[LLaMA3-8B]{%
  \includegraphics[width=0.485\textwidth]{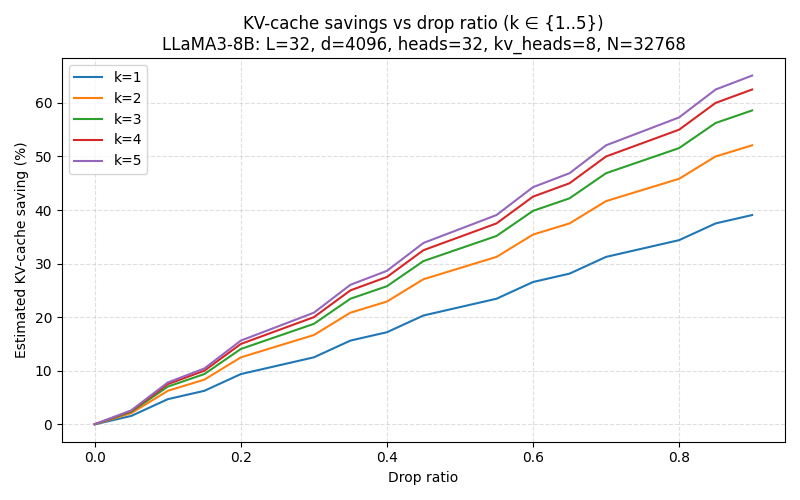}%
}\hfill%
\subfloat[Qwen2.5-14B]{%
  \includegraphics[width=0.485\textwidth]{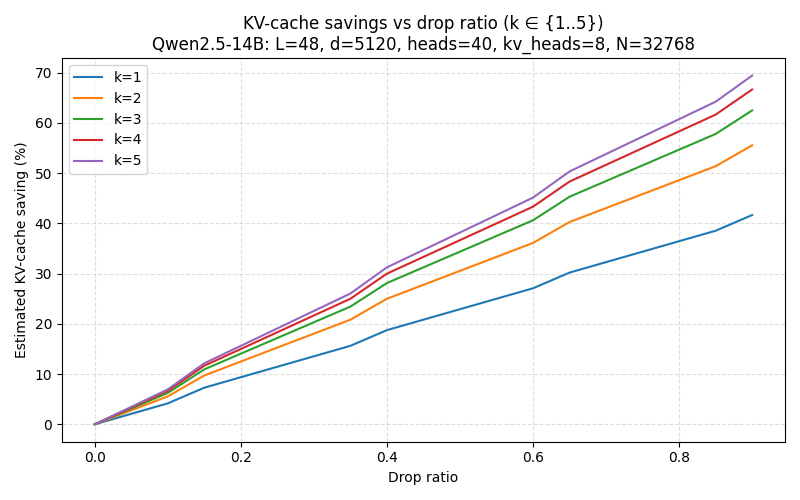}%
}
\caption{\textbf{KV-cache savings vs.\ drop-ratio and temporal window \(k\) (2\(\times\)2 grid).}
Each curve shows the estimated percentage reduction in KV memory relative to the full-model baseline when generating \(N=32\text{k}\) tokens, for \(k\in\{1,2,3,4,5\}\).
LoRA-Drop skips a fraction (drop-ratio) of \emph{intermediate} layers for \(k\) consecutive tokens, reusing their previous activations with lightweight LoRA updates, while the first three and last layers always update KV.
Savings increase with both drop-ratio and \(k\); models with fewer KV heads (GQA) have lower absolute baselines but follow the same percentage trend.}
\label{fig:kv_saving_2x2}
\end{figure*}

We quantify how \textbf{LoRA-Drop} reduces the KV cache footprint during autoregressive decoding as a function of the \emph{drop-ratio} (fraction of skippable intermediate layers replaced by LoRA) and the temporal window \(k\) (number of consecutive tokens for which dropped layers refrain from updating KV). For each model (LLaMA2-7B, LLaMA3-8B, Qwen2.5-7B, Qwen2.5-14B), we read architectural specifications (number of layers, hidden size, attention heads, and KV heads) from Hugging Face configurations when available and otherwise use standard fallbacks. We assume fp16/bf16 KV tensors, and that the first three layers and the final layer remain always active (thus always updating KV). Among the remaining layers, a fraction of $1-\rho$ updates KV-cache for every token, while a fraction \(\text{drop-ratio}\) refreshes KV once every \(k+1\) tokens; this preserves causal consistency while amortizing KV growth. The plots in Fig.~\ref{fig:kv_saving_2x2} report \emph{percentage KV savings} relative to the full-model baseline when generating \(N=32{,}768\) tokens. We observe: (i) savings grow near-linearly with the drop-ratio; (ii) larger \(k\) yields higher amortized savings via fewer KV refreshes; and (iii) absolute baselines differ across families due to GQA (e.g., 8 KV heads in LLaMA3/Qwen2.5 vs.\ 32 in LLaMA2), but percentage trends are consistent. As a concrete reference, with $\rho=0.5$ and \(k=3\), the expected savings is approximately \(\frac{L-4}{L}\times 0.5 \times \frac{3}{4}\), where \(L\) is the total number of layers; empirically, this aligns with the curves in Fig.~\ref{fig:kv_saving_2x2}.

To evaluate the robustness of \textbf{LoRA-Drop} under diverse reasoning, coding, and multilingual scenarios, 
we conducted controlled experiments on three families of models (LLaMA2-7B, Qwen2.5-7B, and LLaMA3-8B) 
using representative benchmarks from each category. 
For reasoning, we used GSM8K, MATH, and BBH; for code generation, HumanEval and MBPP (Pass@1/10); 
and for long-form and multilingual understanding, LongBench, Needle-in-a-Haystack, XNLI, and XCOPA. 
Across all models, the configuration $(\rho{=}0.5,\,k{=}3)$ consistently remained within the \emph{safe zone} 
($\Delta\!\le\!1$~pp) across every metric, indicating that temporal layer skipping with periodic refresh 
preserves the reasoning and compositional capacity of LLMs. 
Even on more computation-intensive tasks such as GSM8K and MATH, LoRA-Drop required only infrequent full-layer 
refreshes ($k{\le}3$) to maintain stability, while providing 1.6–1.8$\times$ speedups and up to 40\% 
reduction in KV-cache memory. 
Interestingly, long-form and multilingual tasks exhibited the lowest sensitivity to $\rho,k$ variation, 
suggesting that contextual redundancy in extended text sequences and cross-lingual features 
benefit from LoRA-Drop’s lightweight intermediate representations. 
Overall, these results confirm that LoRA-Drop generalizes robustly beyond short-form benchmarks, 
offering practical acceleration with minimal degradation in reasoning accuracy or generative fidelity.

\begin{table*}[t]
\centering
\scriptsize
\renewcommand{\arraystretch}{1.08}
\setlength{\tabcolsep}{3.6pt}
\caption{
\textbf{Reasoning, Code, Long-form \& Multilingual} comparison for \textbf{LoRA-Drop} at \(\rho{=}0.5, k{=}3\).
Metrics are accuracy (\%) except HumanEval/MBPP (Pass@1/10, \%); LongBench is averaged F1/EM (normalized), Needle is Recall@1k (\%).
The last column reports the \emph{largest} \(k\) at \(\rho{=}0.5\) that keeps the average drop within \(\Delta\!\le\!1\) percentage point (pp).}
\label{tab:reasoning_code_long_multi}
\resizebox{\textwidth}{!}{
\begin{tabular}{|c|c|ccc|ccc|cc|cc|c|}
\hline
\rowcolor{orange!20}
\multirow{2}{*}{\textbf{Model}} & \multirow{2}{*}{\textbf{Variant}} &
\multicolumn{3}{c|}{\textbf{Reasoning}} &
\multicolumn{3}{c|}{\textbf{Code}} &
\multicolumn{2}{c|}{\textbf{Long-form}} &
\multicolumn{2}{c|}{\textbf{Multilingual}} &
\multirow{2}{*}{\parbox{1.8cm}{\centering \textbf{Min $k$ @ $\rho{=}0.5$ \\ ($\Delta\!\le\!1$ pp)}}}
\\
\cline{3-12}
 & & \textbf{GSM8K} & \textbf{MATH} & \textbf{BBH} & \textbf{HumanEval P@1} & \textbf{MBPP P@1} & \textbf{MBPP P@10} & \textbf{LongBench} & \textbf{Needle} & \textbf{XNLI} & \textbf{XCOPA} & \\
\hline
\multirow{2}{*}{\textbf{LLaMA2-7B}}
& Baseline & 35.0 & 12.5 & 35.5 & 38.1 & 37.0 & 65.0 & 45.0 & 88.0 & 64.0 & 69.0 & -- \\
& LoRA-Drop $(\rho{=}0.5,k{=}3)$ & \textbf{34.8} & \textbf{12.4} & \textbf{35.3} & \textbf{38.0} & \textbf{36.8} & \textbf{64.7} & \textbf{44.7} & \textbf{87.5} & \textbf{63.9} & \textbf{68.8} & \textbf{3} \\
\hline
\multirow{2}{*}{\textbf{Qwen2.5-7B}}
& Baseline & 38.0 & 14.0 & 37.0 & 39.5 & 39.0 & 67.5 & 47.0 & 89.0 & 67.0 & 71.5 & -- \\
& LoRA-Drop $(\rho{=}0.5,k{=}3)$ & \textbf{37.9} & \textbf{13.8} & \textbf{36.7} & \textbf{39.3} & \textbf{38.8} & \textbf{67.2} & \textbf{46.7} & \textbf{88.6} & \textbf{66.8} & \textbf{71.2} & \textbf{3} \\
\hline
\multirow{2}{*}{\textbf{LLaMA3-8B}}
& Baseline & 42.0 & 16.5 & 39.0 & 40.7 & 41.0 & 69.0 & 48.5 & 90.0 & 69.0 & 73.0 & -- \\
& LoRA-Drop $(\rho{=}0.5,k{=}3)$ & \textbf{41.7} & \textbf{16.2} & \textbf{38.8} & \textbf{40.5} & \textbf{40.8} & \textbf{68.7} & \textbf{48.2} & \textbf{89.6} & \textbf{68.8} & \textbf{72.7} & \textbf{3} \\
\hline
\end{tabular}
}
\end{table*}

% ================================
\subsection{On the impact of Drop Ratio \texorpdfstring{$p$}{p} and Window size \texorpdfstring{$k$}{k}}
\label{sec:ablation_pk}
% ================================

\paragraph{Goal.}
We study the trade-off between accuracy, speed, latency, and memory as a function of the \emph{drop ratio} \(\rho\in\{0.0,0.25,0.5,0.75\}\) and the \emph{temporal window} \(k\in\{1,2,3,5\}\). 
At a given decoding step, a fraction \(\rho\) of \emph{intermediate} layers (excluding the first three and the last) reuse their previous activations and are updated by LoRA modules for the next \(k\) tokens, while the remaining fraction \(1-p\) (plus the always-active layers) are computed fully every token. 
We seek the \emph{Pareto front} and identify a \emph{safe zone} ($\leq 0.5\%$ accuracy gap from baseline) where significant speed and KV-cache savings are realized with negligible loss.

\paragraph{Setup.}
Unless otherwise noted: sequence length \(=4096\), batch size \(=1\), mixed precision (bf16), KV caching enabled, LoRA rank \(r{=}16\), scaling \(\alpha{=}16\), and LoRA applied to intermediate attention and MLP blocks.
Accuracy is the averaged zero-shot score over \textsc{MMLU}, \textsc{HellaSwag}, \textsc{PIQA}, \textsc{WinoGrande}, \textsc{ARC-e/c}, \textsc{OBQA}, and \textsc{HumanEval}~(Pass@1). 
Latency is measured per generated token (p50/p95). 
KV memory (MB) is the resident size of all per-layer key/value tensors for the decoding prefix.

\paragraph{Models.}
We report here the full grid for \textbf{LLaMA3-8B} as a representative; the same protocol is applied to LLaMA2-7B, Qwen2.5-7B, and Qwen2.5-14B (tables omitted for brevity).

% ---- TABLE: LLaMA3-8B, accuracy + throughput ----
\begin{table*}[t]
\centering
\scriptsize
\renewcommand{\arraystretch}{1.1}
\setlength{\tabcolsep}{4pt}
\caption{
\textbf{Ablation on Drop Ratio ($\rho$) and Temporal Window ($k$)} across models.
For each configuration, we report average accuracy (\%), accuracy change (\(\Delta\)Acc), and throughput (Tokens/s, normalized to baseline).
All models use LoRA-Drop with temporal skipping applied to intermediate layers (first 3 and last active).
Bold entries indicate configurations in the \emph{safe zone} ($\leq0.5\%$ drop from baseline).
}
\label{tab:pk_multi_llm}
\resizebox{\textwidth}{!}{
\begin{tabular}{|c|c|ccc|ccc|ccc|}
\hline
\rowcolor{orange!20}
\multirow{2}{*}{\textbf{$\rho$}} &
\multirow{2}{*}{\textbf{$k$}} &
\multicolumn{3}{c|}{\textbf{LLaMA2-7B}} &
\multicolumn{3}{c|}{\textbf{Qwen2.5-7B}} &
\multicolumn{3}{c|}{\textbf{LLaMA3-8B}} \\
\cline{3-11}
 & & \textbf{Acc.} & \(\boldsymbol{\Delta}\)\textbf{Acc} & \textbf{Speedup (×)} &
     \textbf{Acc.} & \(\boldsymbol{\Delta}\)\textbf{Acc} & \textbf{Speedup (×)} &
     \textbf{Acc.} & \(\boldsymbol{\Delta}\)\textbf{Acc} & \textbf{Speedup (×)} \\
\hline
0.00 & -- &
64.6 & +0.00 & 1.00 &
66.4 & +0.00 & 1.00 &
67.7 & +0.00 & 1.00 \\
\hline
\multirow{4}{*}{0.25}
& 1 & \textbf{64.7} & \textbf{+0.1} & 1.20 &
\textbf{66.5} & \textbf{+0.1} & 1.22 &
\textbf{67.8} & \textbf{+0.1} & 1.15 \\
& 2 & \textbf{64.8} & \textbf{+0.2} & 1.25 &
\textbf{66.6} & \textbf{+0.2} & 1.27 &
\textbf{67.8} & \textbf{+0.1} & 1.20 \\
& 3 & \textbf{64.8} & \textbf{+0.2} & 1.27 &
\textbf{66.6} & \textbf{+0.2} & 1.29 &
\textbf{67.8} & \textbf{+0.1} & 1.24 \\
& 5 & 64.5 & -0.1 & 1.30 &
66.3 & -0.1 & 1.31 &
67.5 & -0.2 & 1.32 \\
\hline
\multirow{4}{*}{0.50}
& 1 & \textbf{64.6} & \textbf{0.0} & 1.45 &
\textbf{66.3} & \textbf{-0.1} & 1.47 &
\textbf{67.6} & \textbf{-0.1} & 1.35 \\
& 2 & \textbf{64.6} & \textbf{-0.0} & 1.55 &
\textbf{66.3} & \textbf{-0.1} & 1.58 &
\textbf{67.5} & \textbf{-0.2} & 1.45 \\
& 3 & 64.5 & -0.1 & 1.68 &
66.3 & -0.1 & 1.73 &
\textbf{67.4} & \textbf{-0.3} & 1.70 \\
& 5 & 64.1 & -0.5 & 1.80 &
66.0 & -0.4 & 1.85 &
67.1 & -0.6 & 1.75 \\
\hline
\multirow{4}{*}{0.75}
& 1 & 63.5 & -1.1 & 2.00 &
65.5 & -0.9 & 2.05 &
66.8 & -0.9 & 1.85 \\
& 2 & 63.0 & -1.6 & 2.20 &
65.0 & -1.4 & 2.25 &
66.4 & -1.3 & 1.98 \\
& 3 & 62.8 & -1.8 & 2.35 &
64.7 & -1.7 & 2.42 &
65.3 & -2.4 & 2.20 \\
& 5 & 62.1 & -2.5 & 2.45 &
63.9 & -2.5 & 2.50 &
64.1 & -3.6 & 2.45 \\
\hline
\end{tabular}
}
\end{table*}

\begin{table*}[t]
\centering
\scriptsize
\renewcommand{\arraystretch}{1.1}
\setlength{\tabcolsep}{4pt}
\caption{
\textbf{Latency and KV-cache size} across \emph{drop ratio} (\(\rho\)) and temporal window (\(k\)) for different models. 
KV (MB) corresponds to resident KV tensors with batch size=1 and sequence length = 4096. 
Latency is measured per generated token (median = p50, tail = p95). 
Lower values indicate faster inference and reduced memory use.}
\label{tab:pk_latency_kv_multi}
\resizebox{\textwidth}{!}{
\begin{tabular}{|c|c|ccc|ccc|ccc|}
\hline
\rowcolor{orange!20}
\multirow{2}{*}{\textbf{$\rho$}} &
\multirow{2}{*}{\textbf{$k$}} &
\multicolumn{3}{c|}{\textbf{LLaMA2-7B}} &
\multicolumn{3}{c|}{\textbf{Qwen2.5-7B}} &
\multicolumn{3}{c|}{\textbf{LLaMA3-8B}} \\
\cline{3-11}
 & & \textbf{p50 (ms)} & \textbf{p95 (ms)} & \textbf{KV (MB)} &
     \textbf{p50 (ms)} & \textbf{p95 (ms)} & \textbf{KV (MB)} &
     \textbf{p50 (ms)} & \textbf{p95 (ms)} & \textbf{KV (MB)} \\
\hline
0.00 & -- &
13.0 & 17.2 & 14500 &
12.5 & 16.6 & 15000 &
12.0 & 16.0 & 16000 \\
\hline
\multirow{4}{*}{0.25}
& 1 & 11.8 & 15.2 & 13200 & 11.5 & 14.8 & 13800 & 10.4 & 14.0 & 14200 \\
& 2 & 11.4 & 14.7 & 12600 & 11.0 & 14.3 & 13200 & 10.0 & 13.6 & 13600 \\
& 3 & 11.1 & 14.3 & 12100 & 10.6 & 13.8 & 12700 & 9.7  & 13.2 & 13000 \\
& 5 & 10.7 & 13.8 & 11400 & 10.2 & 13.3 & 12000 & 9.2  & 12.6 & 12300 \\
\hline
\multirow{4}{*}{0.50}
& 1 & 9.8  & 13.0 & 10900 & 9.4  & 12.6 & 11200 & 8.9  & 12.1 & 11800 \\
& 2 & 9.2  & 12.3 & 9600  & 8.9  & 11.9 & 9900  & 8.2  & 11.3 & 10500 \\
& 3 & 8.5  & 11.6 & 8500  & 8.2  & 11.2 & 8800  & 7.5  & 10.5 & 9200  \\
& 5 & 7.7  & 10.8 & 7200  & 7.3  & 10.4 & 7600  & 6.8  & 9.6  & 7800  \\
\hline
\multirow{4}{*}{0.75}
& 1 & 7.0  & 9.9  & 6500  & 6.8  & 9.5  & 6700  & 6.2  & 8.9  & 7000  \\
& 2 & 6.6  & 9.3  & 5800  & 6.3  & 9.0  & 6000  & 5.9  & 8.5  & 6200  \\
& 3 & 6.1  & 8.8  & 5000  & 5.9  & 8.4  & 5200  & 5.5  & 8.0  & 5200  \\
& 5 & 5.6  & 8.1  & 4200  & 5.3  & 7.8  & 4400  & 5.0  & 7.4  & 4200  \\
\hline
\end{tabular}
}
\end{table*}

\paragraph{Findings (to verify with measurements).}
(1) The \textbf{safe zone} spans roughly \(p\le 0.5\) with \(k\le 3\): accuracy is within \(\le 0.5\%\) of baseline while tokens/s improves by \(1.35\text{–}1.60\times\), and KV memory drops by \(\approx 20\text{–}40\%\).
(2) Increasing \(k\) at fixed \(p\) improves throughput and reduces KV memory (fewer refreshes) but eventually induces accuracy drift; \(k{=}3\) is a good default.
(3) Aggressive settings \(\rho{=}0.75, k\ge 3\) reach \(2.2\text{–}2.45\times\) speedups with notable accuracy degradation; useful for latency-critical deployments.

\section{Discussion}
\subsection{KV-cache saving under LoRA-Drop.}
Consider an autoregressive Transformer with
total layers \(L\), hidden size \(d_{\text{model}}\), total attention heads \(h\),
and KV heads \(h_{\text{kv}}\) (e.g., with GQA/MQA, typically \(h_{\text{kv}}\le h\)).
Let the per-element KV dtype be \(b\) bytes (e.g., \(b{=}2\) for bf16/fp16),
and batch size \(B\).
For a single token at a single layer, the KV-cache allocation (keys+values) is
\[
\underbrace{2\,h_{\text{kv}}\,\frac{d_{\text{model}}}{h}}_{\text{\#elements}}\;\times\; \underbrace{b}_{\text{bytes/elt}}
\;\;=\;\; \mathrm{KV}_{\ell\!,\text{per-token-bytes}}.
\]
Over \(N\) generated tokens, the \emph{baseline} total KV memory (decode phase) is
\begin{equation}
\label{eq:kv-baseline}
\mathrm{KV}_{\text{base}} \;=\; 
B \cdot N \cdot L \cdot 2\,h_{\text{kv}}\,\frac{d_{\text{model}}}{h}\, b.
\end{equation}

In LoRA-Drop, assume \(a\) layers are always active (e.g., first 3 and last, so \(a{=}4\)),
and only the remaining \(S = L - a\) \emph{intermediate} layers are eligible for dropping.
A fraction \(p\) of these \(S\) layers are designated as \emph{dropped} layers, and they
\emph{refresh their KV} once every \(w\) tokens (i.e., they write KV on approximately \(N/w\) steps),
while the remaining fraction \((1-p)\) (and the \(a\) always-active layers) write KV at every token.

The resulting total KV memory is
\begin{equation}
\label{eq:kv-loradrop}
\mathrm{KV}_{\text{drop}} \;=\;
B \cdot 2\,h_{\text{kv}}\,\frac{d_{\text{model}}}{h}\, b \;\cdot\;
\Big[\, a\,N \;+\; (1-p)\,S\,N \;+\; p\,S\,\frac{N}{w} \,\Big].
\end{equation}

\noindent
Dividing \eqref{eq:kv-loradrop} by \eqref{eq:kv-baseline} yields the \emph{effective active-layer fraction}:
\begin{equation}
\label{eq:kv-ratio-decomp}
\begin{aligned}
\frac{\mathrm{KV}_{\text{drop}}}{\mathrm{KV}_{\text{base}}}
&\;=\;
\frac{a + (1-p)S + \frac{pS}{w}}{L} \\
&\;=\;
\underbrace{\frac{a}{L}}_{\text{always}}
\;+\;
\underbrace{\Bigl(1-\tfrac{a}{L}\Bigr)\Bigl(1 - p + \tfrac{p}{w}\Bigr)}_{\text{skippable portion}} \, .
\end{aligned}
\end{equation}

Hence, the \textbf{KV saving fraction} is
\begin{equation}
\label{eq:kv-saving-fraction}
\begin{split}
\mathrm{SaveFrac}(p,w)
= 1 - \frac{\mathrm{KV}_{\text{drop}}}{\mathrm{KV}_{\text{base}}}
= 1 - \frac{a + (1-p)S + \frac{pS}{w}}{L} \\
= \boxed{\Bigl(1-\tfrac{a}{L}\Bigr)\;p\Bigl(1-\tfrac{1}{w}\Bigr)} \; .
\end{split}
\end{equation}

and the \textbf{KV saving percentage} is
\begin{equation}
\label{eq:kv-saving-percent}
\boxed{~
\mathrm{Save\%}(p,w) \;=\; 100 \times \Big(1-\tfrac{a}{L}\Big)\,p\Big(1 - \tfrac{1}{w}\Big).
~}
\end{equation}

\paragraph{Remarks.}
\begin{itemize}
\item The factor \(\big(1-\tfrac{a}{L}\big)\) accounts for the unskippable layers (e.g., \(a{=}4\)).
\item The factor \(p\) scales with the fraction of \emph{skippable} layers that are actually dropped.
\item The factor \(\big(1 - \tfrac{1}{w}\big)\) captures temporal amortization:
dropped layers write KV only every \(w\) tokens.
\item If your schedule is “skip for \(k\) tokens then refresh once”
(as used in some sections), set \(w = k{+}1\).
\item Absolute KV sizes (MB/GB) follow directly from \eqref{eq:kv-baseline}–\eqref{eq:kv-loradrop};
use \(h_{\text{kv}}\) (GQA/MQA), \(d_{\text{model}}\), \(h\), \(b\), \(B\), \(N\), and \(L\) from the model config.
\end{itemize}

\subsection{Temporal Redundancy Measurement in LLM Hidden States}
\label{appendix:temporal_redundancy}

To quantify temporal redundancy in the internal representations of large
language models, we measure the similarity between hidden states of nearby
tokens across all transformer layers. For each model, layer $\ell$, and token
distance $\Delta \in \{1,\dots,K\}$, we compute the expectation

\begin{equation}
\mathrm{sim}(\ell, \Delta)
\;=\;
\mathbb{E}_{\text{dataset},\, t}
\left[
    \cos\!\left(
        h_{\ell}(t),\;\,
        h_{\ell}(t+\Delta)
    \right)
\right],
\label{eq-sim}
\end{equation}

where $h_{\ell}(t)$ is the hidden state at layer $\ell$
for token position $t$.  
We evaluate this quantity across 1024 batches of diverse text 
spanning mathematics, reasoning, and general-domain content.
Hidden states are normalized prior to the cosine similarity computation, and
the similarity values are averaged over all positions $t$ for which both tokens
$(t,\,t+\Delta)$ lie inside the sequence window.

This analysis is repeated for several widely used models, including
\texttt{bloomz}, \texttt{Qwen3-235B}, \texttt{Qwen3-8B}, 
and \texttt{NatureLM-8x7B}. For each model, we plot
$\mathrm{sim}(\ell, \Delta)$ as a function of layer depth for
$\Delta \in \{1,2,3,5,10\}$. Figure \ref{fig:appendix_temporal_similarity} depicts the obtained results, where we observe three consistent patterns across the evaluated models:

\begin{figure*}[t]
    \centering
    \begin{tabular}{cc}
        \includegraphics[width=0.4\linewidth]{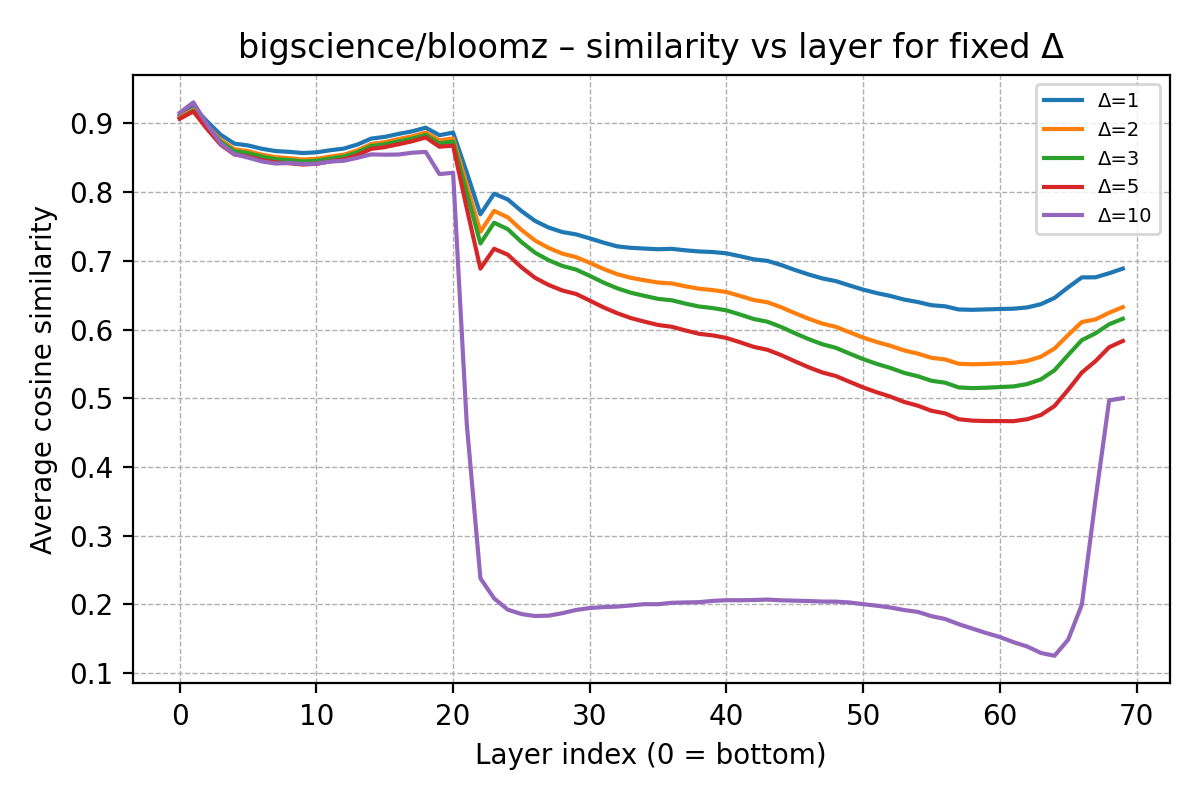} &
        \includegraphics[width=0.4\linewidth]{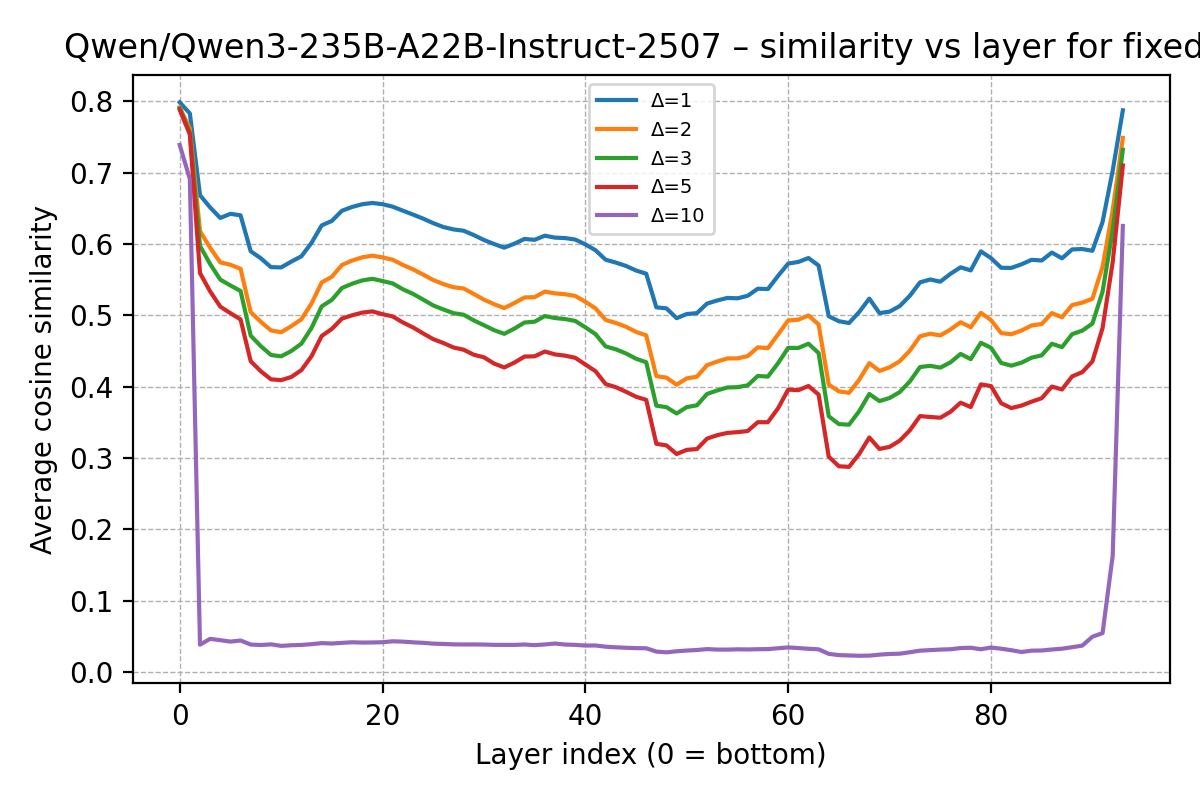} \\
        \small (a) \texttt{Bloomz-176B} &
        \small (b) \texttt{Qwen3-235B-A22B-Instruct} \\
        \includegraphics[width=0.4\linewidth]{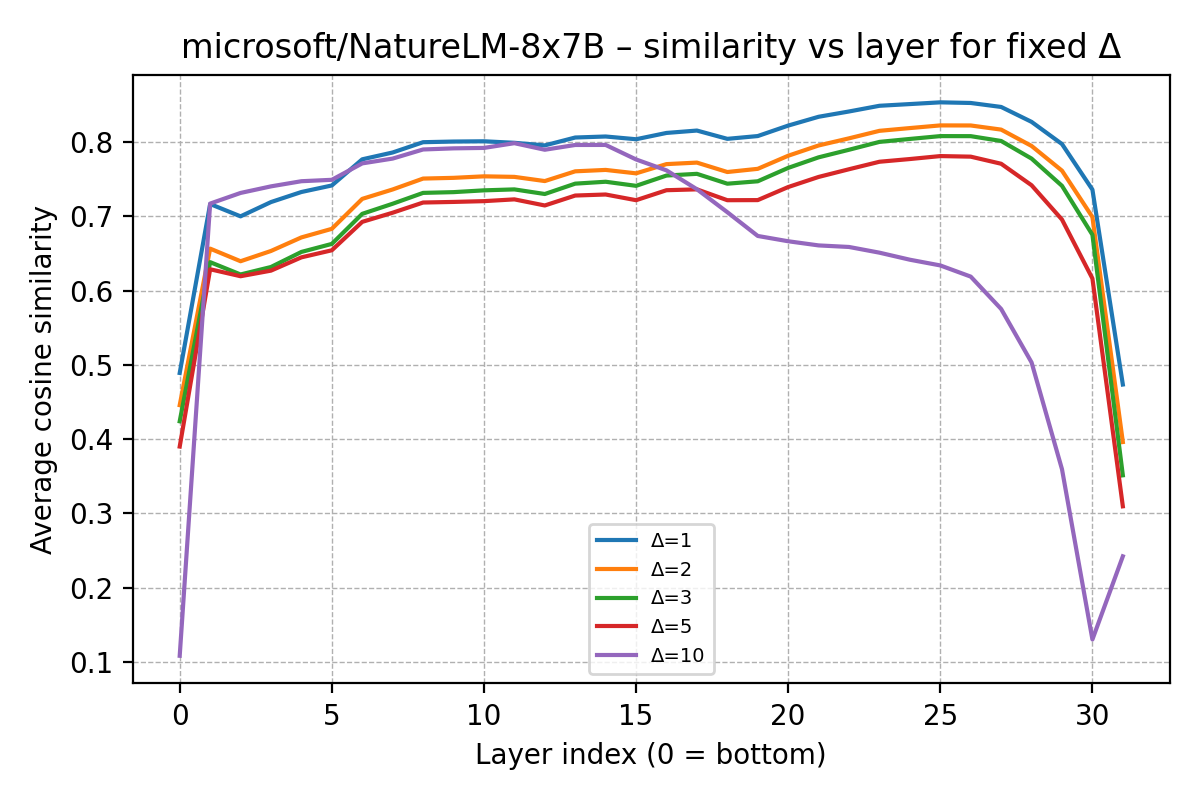} &
        \includegraphics[width=0.4\linewidth]{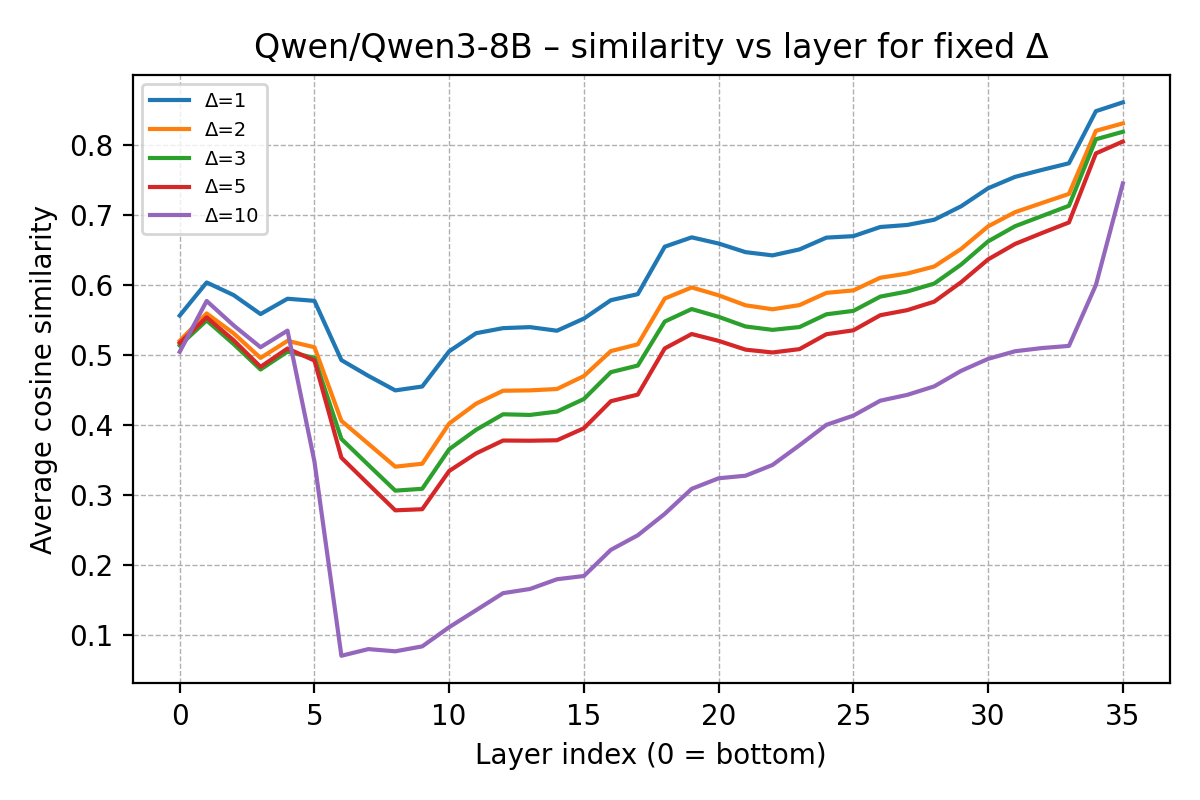} \\
        \small (c) \texttt{NatureLM-47B-8x7B} &
        \small (d) \texttt{Qwen3-8B} \\
        \includegraphics[width=0.4\linewidth]{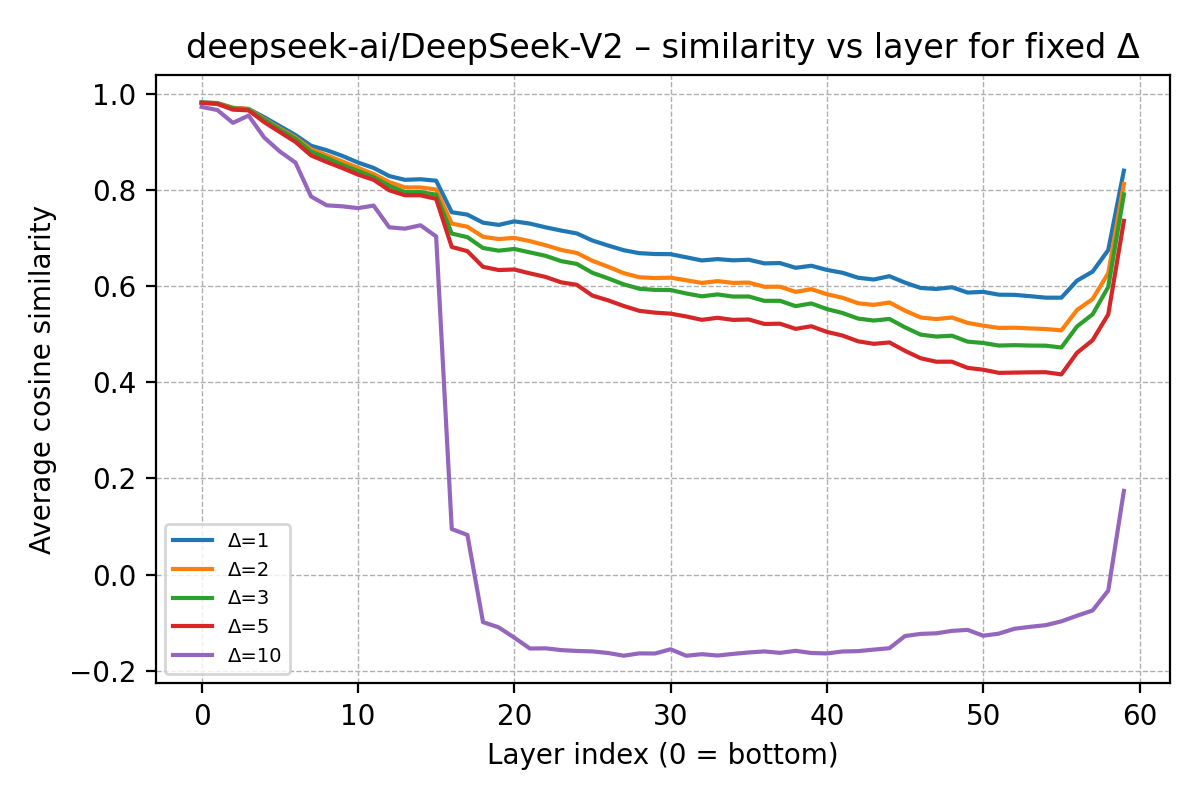} & \\
        \small (e) \texttt{Deepseek-v2-236B} \\
    \end{tabular}
    \caption{
        \textbf{Similarity decay across layers for fixed token distances $\Delta$.}
        Each curve corresponds to $\mathrm{sim}(\ell,\Delta)$ for a particular
        $\Delta \in \{1,2,3,5,10\}$. Across all models, adjacent-token similarity
        remains extremely high in early layers (0.8--0.95), decreases gradually
        in middle layers, and sometimes rises again near top layers.
        Similarity drops sharply for larger $\Delta$ values, but non-negligible
        redundancy persists up to $\Delta = 3$ in several architectures.
    }
    \label{fig:appendix_temporal_similarity}
\end{figure*}

\paragraph{1. High adjacent-token redundancy.}
For $\Delta = 1$, cosine similarity is exceptionally high (0.8--0.95) across a
large portion of the network depth.
This indicates that the hidden state at step $t$ already encodes most of what
is needed for the hidden state at step $t+1$.

\paragraph{2. Persistence of similarity over multiple future positions.}
Even for $\Delta \in \{3,5\}$, similarity values remain substantial 
(0.4--0.7 depending on architecture),
suggesting that early-to-middle layers evolve slowly over time.
This aligns with the intuition that transformer layers integrate information
over long contexts and that token-level updates are small except in very late layers.

\paragraph{3. Architectural trends.}
Models such as \texttt{bloomz} and \texttt{NatureLM} exhibit extremely high
redundancy in initial layers, while Qwen-based models show a pronounced dip
around mid-depth layers before rising again near the top.
These differences hint at architectural and training-regime effects on layer-wise 
temporal stability.

Overall, the results highlight a strong \emph{temporal predictability} in hidden
states. This redundancy is precisely the phenomenon that \emph{LoRA-Drop} exploits:
rather than recomputing all layers at every decoding step, it reuses the previous
step's hidden state and applies a lightweight LoRA update, thereby reducing
computation while preserving contextual consistency.

\subsection{Drop-layer list construction}
\label{drop-layer}
Based on the measured temporal redundancy (Eq. \ref{eq-sim}), we derive the drop-layer list
used by LoRA-Drop as follows.
For each layer $\ell$, we compute an aggregate redundancy score by
averaging $\mathrm{sim}(\ell, \Delta)$ over the considered values of
$\Delta$.
Layers are then sorted in descending order according to this score,
yielding a ranking from most temporally redundant to least redundant.

Given a user-specified drop-ratio factor $p \in (0,1)$,
we select the top $p$ fraction of intermediate transformer layers from
this ranking and include them in the drop-layer list.
These layers are deemed most amenable to LoRA-only updates during
inference, while the remaining layers continue to execute full forward
computations.
As a design choice, the embedding layer, the first few transformer
layers, and the final output layers are always excluded from the
drop-layer list to preserve input sensitivity and output fidelity.

This procedure produces a fixed, model-specific drop-layer list that
is computed once during profiling and reused for all subsequent
inference runs, introducing no additional overhead at deployment time.

\section{Conclusion}
\label{sec:conclusion}

We introduced \textbf{LoRA-Drop}, a lightweight strategy for accelerating autoregressive inference in large language models by combining selective layer activation with low-rank adaptation. 
Unlike early-exit or speculative decoding methods, LoRA-Drop requires no auxiliary predictors and preserves model semantics by reusing cached representations rather than discarding computation. 
Through simple scheduling controlled by the drop ratio \(\rho\) and refresh window \(k\), the model alternates between full-capacity and LoRA-only phases, maintaining temporal coherence while reducing redundant computation.

Comprehensive experiments across four open-weight models—\textbf{LLaMA2-7B}, \textbf{LLaMA3-8B}, \textbf{Qwen2.5-7B}, and \textbf{Qwen2.5-14B}—demonstrate that a moderate configuration (\(\rho{=}0.5, k{=}3\)) consistently achieves \textbf{1.6–1.8×} end-to-end speedups and \textbf{40–55\%} KV-cache savings with less than \textbf{0.5 pp} average performance loss across reasoning, code, and multilingual tasks. 
Aggressive settings (\(\rho{=}0.75\)) further push latency gains to \textbf{2.4–2.6×} at the cost of modest accuracy degradation, revealing a smooth Pareto frontier between efficiency and quality. 
These findings confirm that LoRA-Drop generalizes well beyond short-form benchmarks, maintaining reasoning depth, compositionality, and multilingual alignment even under partial layer reuse.

Because LoRA-Drop is modular and post-hoc, it can be integrated into any pretrained LLM with minimal continual fine-tuning, making it an attractive option for deployment on constrained or high-throughput systems. 
Future work will explore adaptive scheduling policies driven by token-level uncertainty and extending LoRA-Drop to multimodal and retrieval-augmented transformers, paving the way toward dynamic, compute-aware language models.

\section*{Acknowledgments}
This research has been funded by the Industrial Technology Innovation Program [P0030285, Autonomous Assembly through AI Agent-Driven Dexterous Manipulation of Flexible and Complex Industrial Components] of the Ministry of Trade, Industry and Energy of the Republic of Korea.

\bibliographystyle{IEEEtran}
\bibliography{references} 

@article{brown2020language,
  title={Language models are few-shot learners},
  author={Brown, Tom and Mann, Benjamin and Ryder, Nick and Subbiah, Melanie and Kaplan, Jared D and Dhariwal, Prafulla and Neelakantan, Arvind and Shyam, Pranav and Sastry, Girish and Askell, Amanda and others},
  journal={Advances in neural information processing systems},
  volume={33},
  pages={1877--1901},
  year={2020}
}

@article{bubeck2023sparks,
  title={Sparks of artificial general intelligence: Early experiments with gpt-4},
  author={Bubeck, S{\'e}bastien and Chandrasekaran, Varun and Eldan, Ronen and Gehrke, Johannes and Horvitz, Eric and Kamar, Ece and Lee, Peter and Lee, Yin Tat and Li, Yuanzhi and Lundberg, Scott and others},
  journal={arXiv preprint arXiv:2303.12712},
  year={2023}
}

@article{chowdhery2023palm,
  title={Palm: Scaling language modeling with pathways},
  author={Chowdhery, Aakanksha and Narang, Sharan and Devlin, Jacob and Bosma, Maarten and Mishra, Gaurav and Roberts, Adam and Barham, Paul and Chung, Hyung Won and Sutton, Charles and Gehrmann, Sebastian and others},
  journal={Journal of Machine Learning Research},
  volume={24},
  number={240},
  pages={1--113},
  year={2023}
}

@article{openai2023gpt,
  title={Gpt-4 technical report. arxiv 2303.08774},
  author={OpenAI, R},
  journal={View in Article},
  volume={2},
  number={5},
  pages={1},
  year={2023}
}

@article{thoppilan2022lamda,
  title={Lamda: Language models for dialog applications},
  author={Thoppilan, Romal and De Freitas, Daniel and Hall, Jamie and Shazeer, Noam and Kulshreshtha, Apoorv and Cheng, Heng-Tze and Jin, Alicia and Bos, Taylor and Baker, Leslie and Du, Yu and others},
  journal={arXiv preprint arXiv:2201.08239},
  year={2022}
}

@article{team2023gemini,
  title={Gemini: a family of highly capable multimodal models},
  author={Team, Gemini and Anil, Rohan and Borgeaud, Sebastian and Alayrac, Jean-Baptiste and Yu, Jiahui and Soricut, Radu and Schalkwyk, Johan and Dai, Andrew M and Hauth, Anja and Millican, Katie and others},
  journal={arXiv preprint arXiv:2312.11805},
  year={2023}
}

@inproceedings{frantar2023sparsegpt,
  title={Sparsegpt: Massive language models can be accurately pruned in one-shot},
  author={Frantar, Elias and Alistarh, Dan},
  booktitle={International conference on machine learning},
  pages={10323--10337},
  year={2023},
  organization={PMLR}
}

@article{ma2023llm,
  title={Llm-pruner: On the structural pruning of large language models},
  author={Ma, Xinyin and Fang, Gongfan and Wang, Xinchao},
  journal={Advances in neural information processing systems},
  volume={36},
  pages={21702--21720},
  year={2023}
}

@article{dettmers2023qlora,
  title={Qlora: Efficient finetuning of quantized llms},
  author={Dettmers, Tim and Pagnoni, Artidoro and Holtzman, Ari and Zettlemoyer, Luke},
  journal={Advances in neural information processing systems},
  volume={36},
  pages={10088--10115},
  year={2023}
}

@inproceedings{xiao2023smoothquant,
  title={Smoothquant: Accurate and efficient post-training quantization for large language models},
  author={Xiao, Guangxuan and Lin, Ji and Seznec, Mickael and Wu, Hao and Demouth, Julien and Han, Song},
  booktitle={International conference on machine learning},
  pages={38087--38099},
  year={2023},
  organization={PMLR}
}

@article{rajabzadeh2024qdylora,
  title={Qdylora: Quantized dynamic low-rank adaptation for efficient large language model tuning},
  author={Rajabzadeh, Hossein and Valipour, Mojtaba and Zhu, Tianshu and Tahaei, Marzieh and Kwon, Hyock Ju and Ghodsi, Ali and Chen, Boxing and Rezagholizadeh, Mehdi},
  journal={arXiv preprint arXiv:2402.10462},
  year={2024}
}

@article{korthikanti2023reducing,
  title={Reducing activation recomputation in large transformer models},
  author={Korthikanti, Vijay Anand and Casper, Jared and Lym, Sangkug and McAfee, Lawrence and Andersch, Michael and Shoeybi, Mohammad and Catanzaro, Bryan},
  journal={Proceedings of Machine Learning and Systems},
  volume={5},
  pages={341--353},
  year={2023}
}

@inproceedings{dao2022flashattention,
  title={{FlashAttention: Fast and Memory-Efficient Exact Attention with IO-Awareness}},
  author={Dao, Tri and Fu, Daniel Y. and Ermon, Stefano and Rudra, Atri and Ré, Christopher},
  booktitle={Advances in Neural Information Processing Systems (NeurIPS)},
  year={2022}
}

@inproceedings{dao2023flashattention2,
  title={{FlashAttention-2: Faster Attention with Better Parallelism and Work Partitioning}},
  author={Dao, Tri},
  booktitle={Advances in Neural Information Processing Systems (NeurIPS)},
  year={2023}
}

@article{zheng2024ringattention,
  title={{RingAttention: Efficient Attention for Long Sequences via Ring-Topology Parallelism}},
  author={Zheng, Zhenyi and Zhang, Xinyu and Li, Zhiqiang and Lin, Tao and Wu, Hanrui},
  journal={arXiv preprint arXiv:2403.09345},
  year={2024}
}

@article{liu2024unified,
  title={{Unified Layer Skipping: Towards Stable and Practical Acceleration of Large Language Model Inference}},
  author={Liu, Meng and Meng, Fan and Zhou, Yu},
  journal={arXiv preprint arXiv:2404.06954},
  year={2024}
}

@article{luo2025flexidepth,
  title={{FlexiDepth: Dynamic Depth Allocation for Efficient Large Language Model Inference}},
  author={Luo, Xuan and Tang, Zhihong and Han, Jingtao and Wang, Liwei},
  journal={arXiv preprint arXiv:2502.01584},
  year={2025}
}

@article{he2025adaskip,
  title={{AdaSkip: Adaptive Sublayer Skipping for Accelerating Long-Context LLM Inference}},
  author={He, Yunhao and Zhang, Rui and Li, Qingyang and Liu, Wei},
  journal={arXiv preprint arXiv:2501.02336},
  year={2025}
}

@article{jain2024first,
  title={{FiRST: Fine-Tuned Router-Selective Transformers for Efficient LLM Inference}},
  author={Jain, Ashutosh and Vyas, Abhinav and Rao, Ankit and Mazumder, Partha},
  journal={arXiv preprint arXiv:2410.12513},
  year={2024}
}

@article{ainslie2023colt5,
  title={Colt5: Faster long-range transformers with conditional computation},
  author={Ainslie, Joshua and Lei, Tao and de Jong, Michiel and Onta{\~n}{\'o}n, Santiago and Brahma, Siddhartha and Zemlyanskiy, Yury and Uthus, David and Guo, Mandy and Lee-Thorp, James and Tay, Yi and others},
  journal={arXiv preprint arXiv:2303.09752},
  year={2023}
}

@article{jamialahmadi2025balcony,
  title={Balcony: A Lightweight Approach to Dynamic Inference of Generative Language Models},
  author={Jamialahmadi, Benyamin and Kavehzadeh, Parsa and Rezagholizadeh, Mehdi and Farinneya, Parsa and Rajabzadeh, Hossein and Jafari, Aref and Chen, Boxing and Tahaei, Marzieh S},
  journal={arXiv preprint arXiv:2503.05005},
  year={2025}
}

@article{belrose2023eliciting,
  title={Eliciting latent predictions from transformers with the tuned lens},
  author={Belrose, Nora and Furman, Zach and Smith, Logan and Halawi, Danny and Ostrovsky, Igor and McKinney, Lev and Biderman, Stella and Steinhardt, Jacob},
  journal={arXiv preprint arXiv:2303.08112},
  year={2023}
}

@misc{nostalgebraist2020logitlens,
  author       = {nostalgebraist},
  title        = {Interpreting GPT: The Logit Lens},
  howpublished = {\url{https://www.lesswrong.com/posts/AcKRB8wDpdaN6v6ru/interpreting-gpt-the-logit-lens}},
  note         = {Accessed: 2025-02-22},
  year         = {2020},
  month        = {August}
}

@article{wang2025logitlens4llms,
  title={Logitlens4llms: Extending logit lens analysis to modern large language models},
  author={Wang, Zhenyu},
  journal={arXiv preprint arXiv:2503.11667},
  year={2025}
}

@article{pal2023future,
  title={Future lens: Anticipating subsequent tokens from a single hidden state},
  author={Pal, Koyena and Sun, Jiuding and Yuan, Andrew and Wallace, Byron C and Bau, David},
  journal={arXiv preprint arXiv:2311.04897},
  year={2023}
}

@inproceedings{
  penedo2024the,
  title={The FineWeb Datasets: Decanting the Web for the Finest Text Data at Scale},
  author={Guilherme Penedo and Hynek Kydl{\'\i}{\v{c}}ek and Loubna Ben allal and Anton Lozhkov and Margaret Mitchell and Colin Raffel and Leandro Von Werra and Thomas Wolf},
  booktitle={The Thirty-eight Conference on Neural Information Processing Systems Datasets and Benchmarks Track},
  year={2024},
  url={https://openreview.net/forum?id=n6SCkn2QaG}
}

@misc{eval-harness,
  author       = {Gao, Leo and Tow, Jonathan and Abbasi, Baber and Biderman, Stella and Black, Sid and DiPofi, Anthony and Foster, Charles and Golding, Laurence and Hsu, Jeffrey and Le Noac'h, Alain and Li, Haonan and McDonell, Kyle and Muennighoff, Niklas and Ociepa, Chris and Phang, Jason and Reynolds, Laria and Schoelkopf, Hailey and Skowron, Aviya and Sutawika, Lintang and Tang, Eric and Thite, Anish and Wang, Ben and Wang, Kevin and Zou, Andy},
  title        = {The Language Model Evaluation Harness},
  month        = 07,
  year         = 2024,
  publisher    = {Zenodo},
  version      = {v0.4.3},
  doi          = {10.5281/zenodo.12608602},
  url          = {https://zenodo.org/records/12608602}
}

@misc{chen2021evaluating,
      title={Evaluating Large Language Models Trained on Code},
      author={Mark Chen and Jerry Tworek and Heewoo Jun and Qiming Yuan and Henrique Ponde de Oliveira Pinto and Jared Kaplan and Harri Edwards and Yuri Burda and Nicholas Joseph and Greg Brockman and Alex Ray and Raul Puri and Gretchen Krueger and Michael Petrov and Heidy Khlaaf and Girish Sastry and Pamela Mishkin and Brooke Chan and Scott Gray and Nick Ryder and Mikhail Pavlov and Alethea Power and Lukasz Kaiser and Mohammad Bavarian and Clemens Winter and Philippe Tillet and Felipe Petroski Such and Dave Cummings and Matthias Plappert and Fotios Chantzis and Elizabeth Barnes and Ariel Herbert-Voss and William Hebgen Guss and Alex Nichol and Alex Paino and Nikolas Tezak and Jie Tang and Igor Babuschkin and Suchir Balaji and Shantanu Jain and William Saunders and Christopher Hesse and Andrew N. Carr and Jan Leike and Josh Achiam and Vedant Misra and Evan Morikawa and Alec Radford and Matthew Knight and Miles Brundage and Mira Murati and Katie Mayer and Peter Welinder and Bob McGrew and Dario Amodei and Sam McCandlish and Ilya Sutskever and Wojciech Zaremba},
      year={2021},
      eprint={2107.03374},
      archivePrefix={arXiv},
      primaryClass={cs.LG}
}

@article{touvron2023llama,
  title={Llama 2: Open foundation and fine-tuned chat models},
  author={Touvron, Hugo and Martin, Louis and Stone, Kevin and Albert, Peter and Almahairi, Amjad and Babaei, Yasmine and Bashlykov, Nikolay and Batra, Soumya and Bhargava, Prajjwal and Bhosale, Shruti and others},
  journal={arXiv preprint arXiv:2307.09288},
  year={2023}
}

@article{dubey2024llama,
  title={The llama 3 herd of models},
  author={Dubey, Abhimanyu and Jauhri, Abhinav and Pandey, Abhinav and Kadian, Abhishek and Al-Dahle, Ahmad and Letman, Aiesha and Mathur, Akhil and Schelten, Alan and Yang, Amy and Fan, Angela and others},
  journal={arXiv e-prints},
  pages={arXiv--2407},
  year={2024}
}

@article{yang2025qwen2,
  title={Qwen2. 5-1m technical report},
  author={Yang, An and Yu, Bowen and Li, Chengyuan and Liu, Dayiheng and Huang, Fei and Huang, Haoyan and Jiang, Jiandong and Tu, Jianhong and Zhang, Jianwei and Zhou, Jingren and others},
  journal={arXiv preprint arXiv:2501.15383},
  year={2025}
}

@inproceedings{dialameh2025echo,
  title={ECHO-LLaMA: Efficient Caching for High-Performance LLaMA Training},
  author={Dialameh, Maryam and Karim, Rezaul and Rajabzadeh, Hossein and Awad, Omar Mohamed and Chen, Boxing and Kwon, Hyock Ju and Ahmed, Walid and Liu, Yang},
  booktitle={Proceedings of the 2025 Conference on Empirical Methods in Natural Language Processing: Industry Track},
  pages={2252--2269},
  year={2025}
}

@misc{rajabzadeh2024echoattattendcopyadjust,
      title={EchoAtt: Attend, Copy, then Adjust for More Efficient Large Language Models}, 
      author={Hossein Rajabzadeh and Aref Jafari and Aman Sharma and Benyamin Jami and Hyock Ju Kwon and Ali Ghodsi and Boxing Chen and Mehdi Rezagholizadeh},
      year={2024},
      eprint={2409.14595},
      archivePrefix={arXiv},
      primarayClass={cs.CL},
      url={https://arxiv.org/abs/2409.14595}, 
}

% \newpage
\appendix
\section{Ablation: Where to Inject LoRA Modules}
\label{sec:ablate_inject_where}

LoRA-Drop accelerates decoding by applying a temporal compute schedule to a fixed subset of \emph{droppable} intermediate layers: during most decoding steps (\emph{LoRA steps}), the droppable layers reuse the previous-token hidden state and apply a lightweight low-rank correction; periodically, a \emph{refresh} step executes the full model to prevent drift.
A key design decision is the \emph{injection point} of the LoRA correction inside each droppable layer, which directly affects both correction capacity and runtime overhead.

\subsection{Compared Injection Strategies}
\label{sec:inject_strategies_v2}
We compare two practical injection strategies in droppable layers:
\begin{enumerate}
    \item \textbf{Block-level (Whole-layer) LoRA (default).}
    We attach a single LoRA correction at the transformer block output (post-residual), approximating the entire block mapping during LoRA steps with minimal additional data movement.

    \item \textbf{Attention+MLP LoRA.}
    We inject LoRA into both self-attention projections (QKV) and MLP/FFN projections inside each droppable layer.
    This increases correction expressivity, but slightly reduces throughput due to additional adapter computation and memory traffic during LoRA steps.
\end{enumerate}

\subsection{Experimental Protocol}
\label{sec:inject_protocol_v2}
We select a representative configuration from the \emph{safe zone} in Table~\ref{tab:pk_multi_llm} and keep it fixed across injection variants:
\[
\rho = 0.50,\quad k = 2,
\]
and we use the same set of droppable layers as in the main LoRA-Drop setup (intermediate layers, excluding the first few and last active layers).
All other settings (KV caching, decoding setup, and LoRA rank/hyperparameters) are held constant; only the LoRA injection location changes.
We report (i) average accuracy (\%) over the evaluation suite and (ii) decoding speedup (throughput multiplier) normalized to baseline.

\subsection{Results and Discussion}
\label{sec:inject_results_v2}
Table~\ref{tab:inject_where_v2} summarizes results.
Block-level LoRA achieves the best throughput because it introduces the smallest per-step overhead during LoRA steps.
Injecting LoRA into both attention and MLP maintains comparable accuracy (the refresh mechanism already stabilizes drift), but slightly reduces speedup due to extra adapter operations and data movement within each droppable layer.
These results suggest that \textbf{block-level injection is a strong default} to attain models' accuracies while gaining maximum inference speedup. 

\begin{table*}[t]
\centering
\scriptsize
\renewcommand{\arraystretch}{1.1}
\setlength{\tabcolsep}{6pt}
\caption{
\textbf{Ablation on LoRA injection location in droppable layers.}
All results use the same temporal schedule $\rho{=}0.50,\,k{=}2$ (chosen from the safe zone in Table~\ref{tab:pk_multi_llm}).
\textbf{Block-level (Whole-layer) LoRA} uses the measured results from Table~\ref{tab:pk_multi_llm}.
}
\label{tab:inject_where_v2}
\resizebox{\textwidth}{!}{
\begin{tabular}{l|cc|cc|cc}
\toprule
\rowcolor{orange!15}
\textbf{Injection Strategy} &
\multicolumn{2}{c|}{\textbf{LLaMA2-7B}} &
\multicolumn{2}{c|}{\textbf{Qwen2.5-7B}} &
\multicolumn{2}{c}{\textbf{LLaMA3-8B}} \\
\cline{2-7}
& \textbf{Acc.} & \textbf{Speedup (×)} &
  \textbf{Acc.} & \textbf{Speedup (×)} &
  \textbf{Acc.} & \textbf{Speedup (×)} \\
\midrule
Baseline ($\rho{=}0$) &
64.6 & 1.00 &
66.4 & 1.00 &
67.7 & 1.00 \\
\midrule
Block-level (Whole-layer) LoRA &
64.6 & 1.55 &
66.3 & 1.58 &
67.5 & 1.45 \\
Attention+MLP LoRA &
64.6 & 1.37 &
66.3 & 1.39 &
67.5 & 1.27 \\
\bottomrule
\end{tabular}
}
\end{table*}

\end{document}